\documentclass[preprint,12pt]{elsarticle}
\usepackage{graphicx}
\usepackage{subcaption}
\usepackage{amsmath,amssymb}
\usepackage{hyperref}
\usepackage{lineno}
\usepackage{xcolor}
\usepackage{booktabs}
\usepackage{siunitx}
\modulolinenumbers[5]
\usepackage{lmodern}

\journal{Remote Sensing of Environment}

\begin{document}

\begin{frontmatter}


\title{Tree species mapping in Denmark: A comparison of spectral-temporal features with geospatial foundation model embeddings}

\author[inst1]{Alkiviadis Koukos\corref{cor1}}
\ead{akou@dhigroup.com}
\author[inst1]{Spyros Kondylatos}
\ead{spko@dhigroup.com}
\author[inst2]{Thomas Nord-Larsen}
\ead{tnl@ign.ku.dk}
\author[inst1]{Lotte Nyborg}
\ead{ln@dhigroup.com}
\author[inst1]{Christian Tøttrup}
\ead{cto@dhigroup.com}
\author[inst1,inst3]{Kenneth Grogan}
\ead{grogantuan@gmail.com}
\cortext[cor1]{Corresponding author}

\affiliation[inst1]{organization={EO Centre of Excellence, DHI},
           addressline={Agern Alle 5}, 
           city={Hørsholm},
           postcode={2970},
           country={Denmark}}
\affiliation[inst2]{organization={Department of Geoscience and Natural Resource Management, University of Copenhagen},
           addressline={Øster Voldgade 10}, 
           city={Copenhagen},
           postcode={1350},
           country={Denmark}}
\affiliation[inst3]{organization={eometrics},
           addressline={Strandvejen 273C}, 
           city={Charlottenlund},
           postcode={2920},
           country={Denmark}}

\begin{abstract}

Detailed information on tree species distribution is important for sustainable forest management, biodiversity conservation, and climate change mitigation. In this study, we map tree species across Denmark using National Forest Inventory plots and EO data, while evaluating the potential of foundation models for large-scale forest characterization. We compare two alternative input representations for tree species classification: (i) manually engineered spectral-temporal features (STF) derived from multi-temporal Sentinel-1 and Sentinel-2 observations, and (ii) embeddings generated by the EO FMs TESSERA and AlphaEarth. Both representations are complemented with canopy height information. Random forest, XGBoost, and Multi-Layer Perceptron (MLP) classifiers are evaluated for all input representations, with separate assessments for pure and mixed forest stands. The STF-based MLP achieves the highest classification performance, yielding macro F1 scores of 0.843 and 0.653 for pure and mixed stands, respectively.  The MLP trained on TESSERA embeddings delivers competitive performance for pure stands, achieving results within 1.1 percentage points of the best-performing model. TESSERA consistently outperforms STF-based models when fewer than approximately 25\% of training plots are available, demonstrating a substantial advantage under limited training data. Multi-year observations systematically improve classification accuracy relative to single-year inputs, while ablation experiments reveal the complementary contributions of Sentinel-1 backscatter, spectral indices, and canopy height data. The best-performing model is subsequently applied at the national scale to generate a 10 m tree species map of Denmark. Area-adjusted validation indicates an overall map accuracy of 79.9\%. The resulting map, released as an open-access product, is the first high-resolution national tree species map of Denmark and provides a valuable resource for forest monitoring, ecological research, and land management applications.

\end{abstract}


\end{frontmatter}

\section{Introduction}
Forest ecosystems are critically important for several Earth-system functions, including terrestrial biodiversity \citep{brockerhoff2017forest}, carbon storage and fluxes \citep{Pan2011} and local to global water cycles \citep{Ellison2017, Smith2023}. Besides their importance to natural processes, forests also provide significant economic value \citep{Taye2021}, act as important aesthetic and recreational amenities \citep{Panagopoulos2009, Edwards2012}, and hold substantial cultural and spiritual value \citep{MariniGovigli2023, dePater2024}. Although much is already known about the distribution of trees and forest cover \citep{Hansen2013,Brandt2023,Yin2017,FerrerVelasco2022,Reiner2023}, large-scale complementary information on tree species is often lacking \citep{fassnacht2016review, bonannella2022forest}.

Denmark's forests have undergone substantial historical change. Following centuries of decline, forest cover reached a minimum of 2--3\% in the early nineteenth century before recovering to around 15\% today \citep{Fritzboeger1994,NordLarsenEtAl2025}.
Although much of the earlier forest recovery can be attributed to monoculture plantations, the more recent Danish forest policy has shifted toward multifunctional management that supports several key aspects, including climate mitigation, groundwater protection, eutrophication reduction, habitat protection, biodiversity conservation and recreation \citep{MiljoMinisteriet2002, greenTripartite2024}. Recent afforestation initiatives and ambitious forest expansion targets further underscore the growing importance of forest ecosystems in national environmental policy \citep{greenTripartite2024}.

The transition towards more diverse forest management and afforestation practices has increased heterogeneity in species composition, forest structure, and spatial configuration \citep{brockerhoff2017forest}. While this diversity can enhance ecosystem resilience and multifunctionality, it also reduces the suitability of conventional stand-level inventory and mapping approaches. Consequently, forest owners and managers increasingly require fine-scale spatial information on tree species composition to support diverse planning and management activities \citep{fassnacht2016review}.

The Danish Nature Agency (Naturstyrelsen; NST) maintains a stand-level forest dataset that records the dominant tree species within individual forest stands across Denmark \citep{danish_ministry_environment_2026_saba}.
Although spatially explicit, the dataset is incomplete because it primarily focuses on publicly managed forests, omitting large areas of privately owned and commercial forests. 
Furthermore, tree species are recorded at the stand level, limiting its suitability for fine-scale forest monitoring and analysis.

Similarly, National Forest Inventories (NFIs) provide detailed and statistically robust information on forest composition and distribution and form an invaluable source of data for national forest monitoring systems.
In Denmark, the NFI maintains a nationwide network of permanent sample plots with detailed measurements of tree species composition and forest structure \citep{NordLarsenEtAl2025}. 
While these observations are invaluable for monitoring forest resources, they are inherently sparse due to the cost and effort associated with field data collection and therefore cannot provide the wall-to-wall spatial information required for operational forest management.

Remote Sensing (RS) has become an essential complement to field inventories, enabling continuous mapping of forest composition across large areas.
In particular, the Sentinel-1 and Sentinel-2 missions provide freely available observations with high spatial and temporal resolution, combining structural and spectral information.
These datasets have stimulated extensive research on forest species classification, using combinations of optical, radar, hyperspectral, and LiDAR observations \citep{fassnacht2016review, pu2021mapping}. 
At national and regional scales, the integration of NFI reference data with Sentinel observations has proven particularly effective for generating spatially explicit tree species maps \citep{blickensdorfer2024national, immitzer2023tree, abdi2026mapping, freudenberg2025sentinel}.

A central challenge in species-level forest classification remains the representation of temporal variability.
Previous studies have shown that species discrimination benefits strongly from preserving finer-scale temporal dynamics rather than relying solely on single-date observations or highly aggregated image composites \citep{immitzer2019optimal, hemmerling2021mapping, blickensdorfer2024national, Pasquarella2018}. 
Consequently, many approaches derive engineered STF from Sentinel-2 time series, including seasonal composites, statistical summaries, and phenological indicators \citep{grabska2019forest, hoscilo2019mapping, hermosilla2022mapping, wang2022assessing}.
These features are often combined with textural or structural information derived from Sentinel-1 observations \citep{blickensdorfer2024national, lechner2022combination}. 
Similar findings have been reported in related land-cover mapping applications, such as crop classification, where dense time series consistently outperform temporally aggregated representations \citep{Griffiths2019, Blickensdoerfer2022}. 
Collectively, these results highlight the importance of preserving temporal information to capture species-specific phenological signatures.

Such engineered feature representations have traditionally been used as inputs to Machine Learning (ML) models such as Random Forest (RF) and gradient boosting algorithms \citep{mu2025national}. 
Recently, deep learning approaches have demonstrated further improvements by learning representations directly from Sentinel time series \citep{xi2021exploitation, mu2025national, mouret2025tree, huang2023spectral}. 
However, these approaches typically require large volumes of labeled training data and are often developed for specific tasks or regions, potentially limiting their transferability across scales and domains \citep{xiao2025foundation}.

Earth Observation (EO) Foundation Models (FMs) have emerged as a promising alternative.
Pre-trained on large-scale multi-modal EO archives, these models learn general-purpose representations that encode spatial, temporal, and semantic information and can be readily transferred to downstream applications \citep{xiao2025foundation}.
Recent models, such as AlphaEarth \citep{brown_alphaearth_2025} and TESSERA \citep{feng_tessera_2025}, have further advanced this paradigm by providing analysis-ready embeddings that encapsulate global EO information in unified pixel-level representations, enabling users to leverage large-scale pre-training without the computational burden of developing task-specific deep learning pipelines. 

Evaluations of such models have so far focused on well-curated benchmark datasets and comparatively coarse-grained tasks \citep{xiao2025foundation, huo2025remote}. 
Only a limited number of studies have explored their potential for fine-grained ecological applications such as tree species mapping \citep{wang2026evaluating, ball2026geospatial, bountos2025fomo}.
While these studies often report encouraging results, comparisons have generally been made against relatively simple baselines, such as annual or seasonal Sentinel composites. 
It therefore remains unclear whether foundation-model embeddings can match or surpass the richer STF representations currently used in operational tree species mapping, especially those designed to preserve detailed phenological dynamics.

In this study, we develop an ML framework for national-scale dominant tree species mapping in Denmark using EO data and NFI plots as reference data. 
We further investigate whether EO FM embeddings can provide a viable alternative to manually engineered features for national-scale tree species classification by comparing models trained on (i) STF extracted from multi-temporal Sentinel-1 and Sentinel-2 observations and (ii) embeddings generated by the EO FMs TESSERA and AlphaEarth. 
To ensure a fair comparison, both representations are combined with canopy height information and evaluated using RF, XGBoost, and Multi-Layer Perceptron (MLP) classifiers. 

The use of NFI data also introduces an important evaluation challenge.
Because NFI plots are designed to characterize forest conditions rather than provide spectrally pure training samples, they frequently contain multiple tree species in varying proportions. 
Consequently, mixed-species stands are common and inherently more challenging to classify than pure stands owing to their greater structural and spectral complexity.
Despite their prevalence in operational forest inventories, mixed stands have received comparatively limited attention in the literature, where evaluations are often restricted to pure stands. 
To address this gap, we assess model performance separately for pure and mixed forest stands, providing a more realistic evaluation under operational forest conditions.

Our main contributions are as follows:
\begin{itemize}
    \item We develop an ML framework for tree species mapping across Denmark, integrating Sentinel-1 and Sentinel-2 time series and metrics with canopy height and the Danish NFI.
    \item We provide a systematic comparison of engineered STF and embeddings derived from FMs, assessing their relative performance under varying training data availability.
    \item We conduct an analysis to identify the key drivers of classification performance, including the contribution of spectral indices, Sentinel-1 data, canopy height data, and the added value of multi-year over single-year data acquisitions.
    \item We release the first open-access, high-resolution national tree species map of Denmark at 10\,m spatial resolution, validated using a sample of both pure and mixed tree cover, providing a valuable resource for forest monitoring, biodiversity assessment, and ecosystem management at the national scale.
\end{itemize}

\section{Data}
\label{sec:data}

\subsection{Study Area}
\label{sec:study_area}

The study area includes forests in Denmark, covering a range of forest types and management conditions. 
Danish forests are generally small and fragmented and often interspersed with agricultural land, urban areas, and infrastructure \citep{johannsen2019danish}. 
This fragmented landscape poses challenges for satellite-based forest mapping and thus provides a suitable context for evaluating tree species classification approaches with ML.

Danish forests are almost evenly split between broadleaved and coniferous tree species, while broadleaved species are increasing due to recent management trends \citep{NordLarsenEtAl2025}. The most common conifers are Norway spruce (\textit{Picea abies} L.) and Sitka spruce (\textit{Picea sitchensis} Bong Carr.), whereas European beech (\textit{Fagus sylvatica} L.) and pedunculate oak (\textit{Quercus robur} L.) are the dominant broadleaved species.
The composition of species varies between regions and reflects differences in site conditions and forest management practices. 
A substantial proportion of the Danish forest consists of mixed species rather than pure stands, reflecting a shift toward more structurally and compositionally diverse forests \citep{NordLarsenEtAl2025}. 
The relatively flat terrain and temperate climate support clear seasonal vegetation dynamics, which can be captured using multi-temporal satellite observations.

\subsection{National Forest Inventory Data}
\label{sec:ndfi}
\begin{figure}[!ht]
    \centering
    \includegraphics[width=1\linewidth]{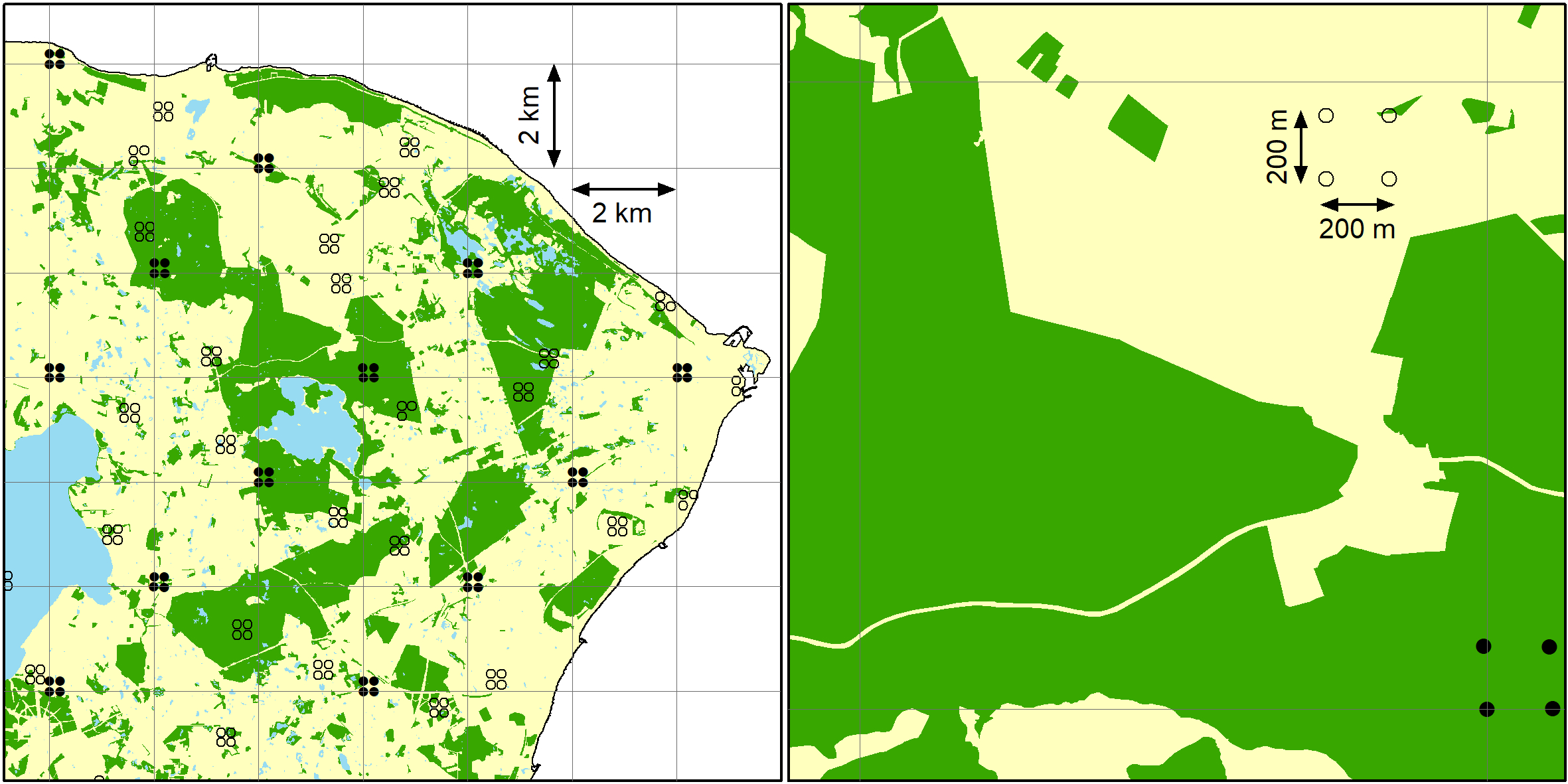}
    \caption{Overview of the Danish NFI sampling design. The left panel shows the systematic $2\,\mathrm{km} \times 2\,\mathrm{km}$ grid over North Zealand, while the right panel provides a detailed view of a single grid cell, illustrating the arrangement of permanent (filled circles) and temporary (open circles) sample clusters.}
    
    \label{fig:nfi_sample_design}
\end{figure}

Reference data are obtained from the Danish NFI, providing species-level labels for model training and validation.
The NFI is organized using a nationwide systematic grid of $2 \times 2$ km cells \citep{NordLarsenJohannsen2016}. 
Within each cell, a cluster of four sample plots is positioned at the corners of a $200 \times 200$ m square (Figure~\ref{fig:nfi_sample_design}). 
Data collection is carried out over a five-year cycle during which one-fifth of the plots— evenly distributed throughout the country — are measured each year.

\begin{figure*}[!ht]

\centering
\includegraphics[width=\linewidth]{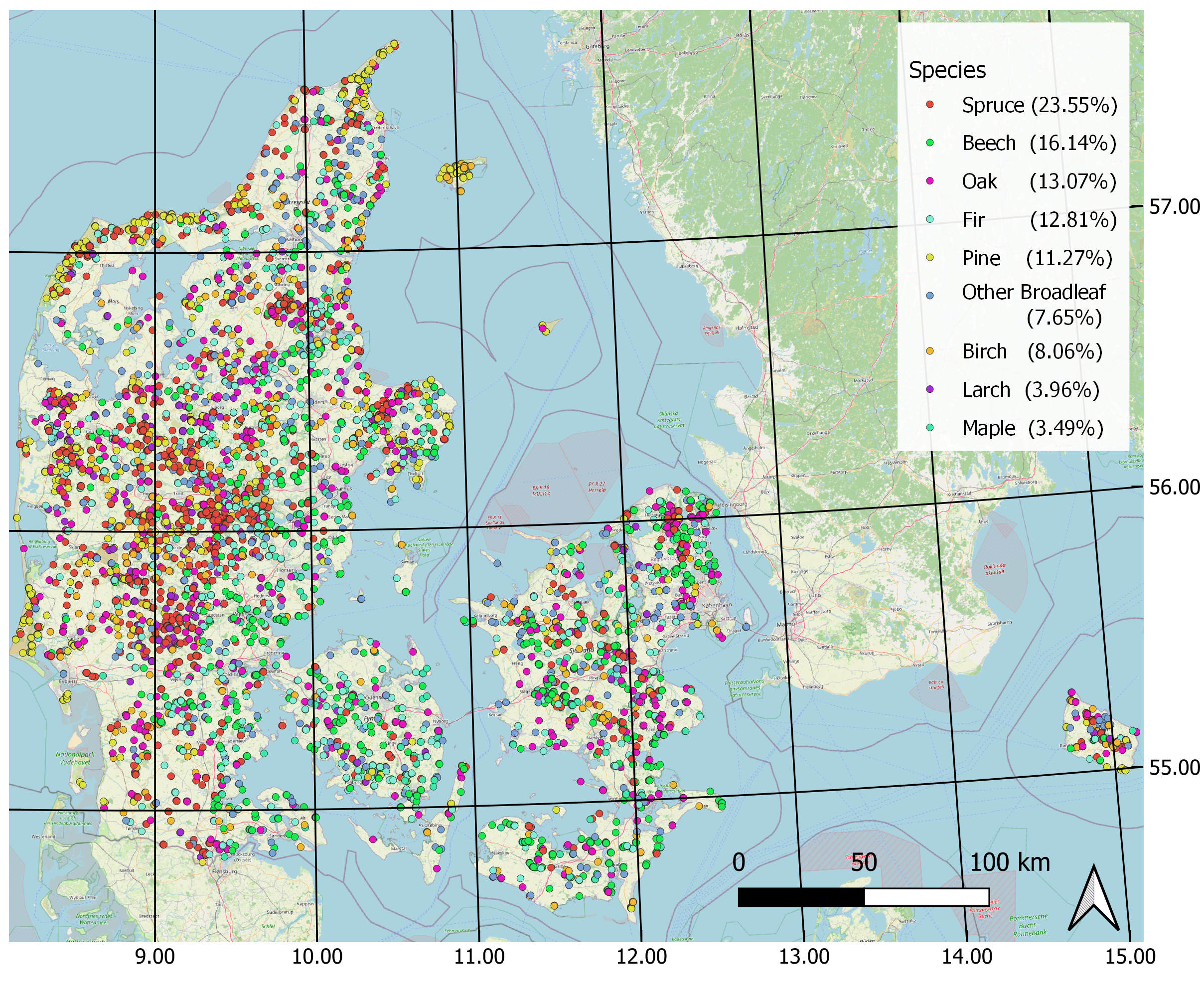}
\caption{Study area and distribution of NFI plots. Colors indicate the dominant tree species at each plot, and percentages in the legend denote the relative proportion of each species class in the dataset.}
\label{fig:study_area}

\end{figure*}

The sampling network consists of permanent and temporary clusters. 
One-third of the clusters are permanent and fixed in the southwest corner of each grid cell; these are remeasured in every five-year cycle. 
The remaining two-thirds are temporary clusters, which are randomly relocated within their respective grid cells for each new cycle. 
The permanent clusters play an important role in tracking the temporal development of the forests.

In total, the network includes around 43,000 plots, although only clusters with at least one sample plot containing forest are measured in the field during a given cycle. 
For the forest classification, the NFI applies the FAO definition of forest \citep{FAO_2020}. 
The forest plots to be measured are identified prior to fieldwork using recent aerial images, typically less than one year old.

\subsubsection{Field data collection}

In the field, the center of each plot is geolocated with high precision, enabling accurate remeasurement and integration with other geographic data sources. 
The circular sample plots are partitioned into subplots wherever they intersect multiple land-use categories or forest stands, ensuring that each subplot represents a homogeneous unit with respect to land use and stand characteristics. 
Further, stand-level recordings include likely origin and management of forest stands as well as previous damage and forest management operations such as thinning, harvest, soil preparation, and planting.

Within the sample plots, all trees taller than 1.3 m are registered within a circle of 3.5 m radius, trees with a diameter at breast height (dbh) greater than 10 cm are registered within a circle of 10\,m radius, and trees with dbh greater than 40 cm are registered within the full 15 m radius circle \citep{NordLarsenJohannsen2016}. 
For each tree, the tree species is recorded, and the diameter is measured at breast height by a single caliper measurement. 
Multiple stems from the same root are considered individual trees. Total tree height is recorded for a subsample of up to six trees in each plot. Additional tree attributes include canopy layer (i.e., which canopy layer the tree belongs to), possible damage, defoliation, and micro-habitats.

\subsubsection{Plot selection and species-group labels}

The Danish NFI contains observations dating back to 2002. 
However, for this study, plots measured between 2014 and 2022 are used, resulting in a total of 9,304 unique plots, and after removing instances of very rare classes, we end with 8,888 plots. 
Since the RS features used in this study may include observations from years following the NFI measurement, it is necessary to account for potential forest disturbances that may have occurred after the reference year.
To address this, we use the European Forest Disturbance Atlas dataset \citep{viana2025european}.
Plots intersecting mapped disturbance events after their field measurement year and before the end of 2022 are removed. 
In total, 225 plots are removed, resulting in a set of 8,663 plots.

In this study, our goal is to identify the dominant tree species present in the upper canopy layer using RS data. 
This information can be derived from the NFI plot measurements. 
Although the dataset contains records for 59 different tree species, most of them occur rarely.
Furthermore, several species are spectrally very similar (e.g., different fir species) and cannot be reliably distinguished using Sentinel satellite data. 
For this reason, species are aggregated into broader species groups. 
This results in nine classification labels: beech (\textit{Fagus sylvatica} L.), birch (mainly \textit{Betula pendula} Roth and \textit{Betula pubescens} Ehrh.), fir (mainly \textit{Abies alba} Mill., \textit{Abies nordmanniana} (Steven) Spach, and \textit{Abies grandis} (Douglas ex D.Don) Lindl.), larch (mainly \textit{Larix decidua} Mill., \textit{Larix kaempferi} (Lamb.) Carrière, and \textit{Larix × eurolepis} A.Henry), maple (mainly \textit{Acer pseudoplatanus} L.), oak (mainly \textit{Quercus robur} L. and \textit{Quercus petraea} (Matt.) Liebl.), pine (mainly \textit{Pinus sylvestris} L., \textit{Pinus contorta} Douglas ex Loudon, and \textit{Pinus mugo} Turra), spruce (mainly \textit{Picea abies} (L.) H.Karst. and \textit{Picea sitchensis} (Bong.) Carrière), and other broadleaves (hereafter OBL).

Using the division of the circular sample plots into sub-plots identified as forest or other land cover, we get the fraction of forest cover ($forest_{frac}$) and the fraction of the dominant forest stand ($stand_{frac}$) for each sample plot. 
For each tree measured, we calculate the basal area ($g=\pi/4 \cdot dbh^2$, where $dbh$ is the diameter measured at breast height) and scale it to the full plot area according to which of the three circles (i.e., 3.5, 10, or 15 m) the tree is measured within. 
The fraction of scaled basal area for each tree species group serves as a proxy for tree cover, owing to the often close relationship between basal area and tree crown size. 
The proportion of the dominant tree species in the upper canopy layer is termed ($dom_{frac}$). 
Based on these attributes, plots are classified as pure forest when $dom_{frac} \geq 0.90$ and $stand_{frac} \geq 0.80$. 
Plots are classified as mixed forest when the product $dom_{frac} \times stand_{frac}$, with both variables expressed as fractions between 0 and 1, is at least 0.50. 
This criterion ensures that the dominant species group represents at least half of the full plot area when accounting for both stand purity and within-stand dominance. 
All remaining plots are considered unsuitable for tree species classification and are discarded from the analysis. Consequently, we retained 2,683 pure plots and 3,098 mixed plots.

\subsection{Additional forest layers}
\label{sec:add_layers}

We map tree species for all pixels covered by the Danish Digital Forest Map \citep{SGAV2025DigitalForestMap}. The map is created for the year 2022, using Sentinel seasonal composites and phenology metrics and has a high reported accuracy of $\sim$98\%. This forest extent map is aligned with the same forest definition used by the Danish NFI. The forest extent map is originally in polygon form and measures $\sim$653,250 ha. This polygon dataset is rasterized to match the 10 m grid used in this study, resulting in a slight reduction in forest area to 651,855 ha.

For visual assessment of our resulting map, we use a polygon-based dataset published by the NST \citep{danish_ministry_environment_2026_saba}. This dataset records the dominant tree species at the broader stand-level for publicly managed forests, missing large areas of privately owned and commercial forests. Nonetheless, it is a useful dataset for checking that broad patterns are captured in our tree species map.

\subsection{Sentinel-2}
\label{sec:sentinel_2}

Sentinel-2 reflectance data are processed to derive time series features for forest classification. 
All available imagery from 2020 to 2022 is used. 
Level-1C products are downloaded from Copernicus data catalogs and processed per tile using the Framework for Operational Radiometric Correction for Environmental monitoring (FORCE) version 3.8.01 \citep{Frantz2019}.
The processing steps include radiative transfer-based atmospheric correction \citep{Frantz2016a}, adjacency effect correction, bi-directional reflectance function correction \citep{Roy2017}, and topographic correction using enhanced C correction \citep{Buchner2020}. 
We use the freely available Danish Digital Elevation Model from the Danish Climate Data Agency as input for topographic correction \citep{DanishClimateDataAgencyDHM}. 
Sentinel-2 imagery is coregistered to ensure per-pixel temporal consistency, and cloud shadow masking is done using Fmask version 4.6 \citep{Zhu2015, Qui2019, Frantz2018}. 
All 20 m bands are resampled to match the resolution and grid of the 10\,m bands.

Following per-tile processing, all imagery and quality data are reprojected to EPSG:25832 and divided into non-overlapping tiles of $2048 \times 2048$ pixels at 10\,m spatial resolution, forming a data cube \citep{giuliani2017building} covering Denmark. 
These data and corresponding metadata are then stored in a Spatio-Temporal Asset Catalog (STAC) for efficient querying, image viewing, retrieval, and higher-level processing, including spectral index calculations, time series interpolation, and spectral temporal metrics \citep{stac_spec}.

All Sentinel-2 bands except 1, 9, and 10 are used as classification features. 
Additionally, a set of spectral indices is derived, selected to capture differences in canopy phenology, moisture content, and chlorophyll-related properties relevant to tree species discrimination \citep{fassnacht2016review, grabska2020evaluation}. 
Based on their relevance for canopy phenology, moisture status, and chlorophyll-related properties, the following indices are used for classification:

\begin{equation}
\mathrm{NDVI} = \frac{B_{08} - B_{04}}{B_{08} + B_{04}}
\end{equation}

\begin{equation}
\mathrm{NDMI} = \frac{B_{8A} - B_{11}}{B_{8A} + B_{11}}
\end{equation}

\begin{equation}
\mathrm{NBR} = \frac{B_{8A} - B_{12}}{B_{8A} + B_{12}}
\end{equation}

\begin{equation}
\mathrm{SWIR}_{\mathrm{ratio}} = \frac{B_{11}}{B_{12}}
\end{equation}

\begin{equation}
\mathrm{LCI} = \frac{B_{08} - B_{05}}{B_{08} + B_{04}}
\end{equation}

\begin{equation}
\begin{aligned}
\mathrm{GEMI} &= \eta (1 - 0.25\eta) - \frac{B_{04} - 0.125}{1 - B_{04}} \\
\eta &= \frac{2(B_{08}^2 - B_{04}^2) + 1.5B_{08} + 0.5B_{04}}{B_{08} + B_{04} + 0.5}
\end{aligned}
\end{equation}

\begin{equation}
\mathrm{ReNDVI} = \frac{B_{08} - B_{07}}{B_{08} + B_{07}}
\end{equation}

\begin{equation}
\mathrm{REPI} = 700 + 40 \cdot \frac{\left(\frac{B_{04} + B_{07}}{2} - B_{05}\right)}{B_{06} - B_{05}}
\end{equation}

To ensure regular temporal intervals and reduce noise in the time series, both spectral bands and derived indices are temporally interpolated using Whittaker-Eilers smoothing \citep{eilers2003perfect}. 
Interpolated time series are generated at 10-day temporal resolution.
In addition, for each year, seasonal summary statistics including maximum, mean, median, standard deviation, and 5th, 25th, 75th, and 95th percentiles are computed.
The interpolated time series and the seasonal statistics are then integrated into the shared datacube infrastructure.

\subsection{Sentinel-1}
\label{sec:sentinel_1}

Sentinel-1 C-band Synthetic Aperture Radar (SAR) data are used to complement the optical Sentinel-2 time series with information primarily related to forest structure. 
All available Sentinel-1 imagery covering Denmark from 2020 to 2022 is used. 
Specifically, Ground Range Detected (GRD) imagery from both ascending and descending orbits, recorded in Interferometric Wide Swath mode, is downloaded from Copernicus. 
Backscatter coefficients in VV and VH polarizations are processed using SNAP \citep{zuhlke2015snap}. 
Each Sentinel-1 image is first calibrated to linear units, followed by speckle reduction using a refined Lee filter. 
The data are then terrain-corrected and converted from linear to decibel units to create a multi-temporal backscatter time series.
In addition, the difference between VH and VV backscatter ($\mathrm{VH} - \mathrm{VV}$) is computed to capture relative changes between the two polarisations, which are sensitive to vegetation structure and surface scattering mechanisms.

\begin{table}[!htbp]
\centering
\footnotesize
\renewcommand{\arraystretch}{1.0}
\caption{List of the inputs and temporal resolutions used to construct the STF set and the FM representations. Sentinel-2 inputs comprise 10-day interpolated time series and seasonal summary statistics, Sentinel-1 inputs comprise monthly summary statistics, FM embeddings are annual, and canopy-height metrics are static.}
\begin{tabular}{ll}
\toprule
\textbf{Feature} & \textbf{Temp. resolution} \\
\midrule

\multicolumn{2}{c}{\textbf{Sentinel-2}} \\
\midrule

Blue (B2) & 10-day + seasonal \\
Green (B3) & 10-day + seasonal \\
Red (B4) & 10-day + seasonal \\
Red-edge (B5–B7) & 10-day + seasonal \\
Near-infrared (B8, B8A) & 10-day + seasonal \\
Short-wave infrared (B11, B12) & 10-day + seasonal \\
NDVI (Norm. Difference Vegetation Index) & 10-day + seasonal \\
NDMI (Norm. Difference Moisture Index) & 10-day + seasonal \\
NBR (Norm. Burn Ratio) & 10-day + seasonal \\
SWIR$_{\text{ratio}}$ (Shortwave Infrared Ratio) & 10-day + seasonal \\
LCI (Leaf Chlorophyll Index) & 10-day + seasonal \\
GEMI (Global Env. Monitoring Index) & 10-day + seasonal \\
ReNDVI (Red-edge NDVI) & 10-day + seasonal \\
REPI (Red-edge Position Index) & 10-day + seasonal \\

\addlinespace
\midrule
\multicolumn{2}{c}{\textbf{Sentinel-1}} \\
\midrule

VV & monthly \\
VH & monthly \\
VH--VV & monthly \\

\addlinespace
\midrule
\multicolumn{2}{c}{\textbf{Foundation Model Embeddings}} \\
\midrule

AlphaEarth & yearly \\
TESSERA & yearly \\

\addlinespace
\midrule
\multicolumn{2}{c}{\textbf{Canopy Height (CH) Data}} \\
\midrule

median CH & static \\
maximum CH & static \\
standard deviation CH & static \\
90th percentile CH & static \\
95th percentile CH & static \\
98th percentile CH & static \\
99th percentile CH & static \\

\bottomrule
\end{tabular}
\end{table}

After preprocessing, the Sentinel-1 data are added to the datacube by reprojecting and retiling the data to match the dimensions, 10\,m grid, and EPSG projection as the Sentinel-2 data. The VH, VV, and VH--VV bands and their metadata are added to the STAC for efficient downstream data requests and processing.

Similar to \cite{blickensdorfer2024national}, Sentinel-1 observations are aggregated at a monthly temporal resolution. For each month, summary statistics including the mean, median, standard deviation, and the 25th and 75th percentiles are computed for VV, VH, and the VH--VV difference. 
These monthly statistics are also integrated into the shared datacube infrastructure.

\subsection{Canopy Height Metrics}
\label{sec:canopy_height}

We use the freely available Digital Terrain Model (DTM) and Digital Surface Model (DSM) components of the Danish Elevation Model, provided by the Danish Climate Data Agency \citep{DanishClimateDataAgencyDHM}. 
Both components are available at a spatial resolution of 0.4~m.
Canopy height is estimated by subtracting the DTM from the DSM, resulting in a canopy height model representing vegetation height above ground. 
The high spatial resolution of the elevation data allows detailed characterization of forest vertical structure.

To integrate canopy height information with the satellite-based features, spatial summary statistics are computed over the 10~m analysis grid. 
These include the median, maximum, standard deviation, and upper percentiles (90th, 95th, 98th, and 99th percentiles). 
The resulting canopy height metrics are used as additional input for the classification models.

\subsection{Foundation Model Embeddings}
\label{sec:fm_embeddings}

FM embeddings provide dense, analysis-ready representations that can serve as an alternative to task-specific feature engineering as inputs to ML classifiers \citep{zhu_foundations_2026}.
In this study, we evaluate the utility of pre-trained embeddings extracted from two geospatial FMs, AlphaEarth \citep{brown_alphaearth_2025} and TESSERA \citep{feng_tessera_2025}.
Annual embeddings from both models are retrieved for years 2020, 2021, and 2022, consistent with the temporal range used for the manually engineered features.
The embeddings are used without any fine-tuning or domain-specific adaptation, enabling an evaluation of their out-of-the-box representational capacity for forest species mapping.
This experimental setting reflects a real-world operational scenario in which FM embeddings can be deployed without any task-specific adjustment.
This provides a use case of growing relevance as pre-trained geospatial models become increasingly accessible to the RS community \citep{janowicz_geofm}. 
The key properties of the embeddings used in this study are summarized in Table~\ref{tab:fms}.

\begin{table}[!ht]
\centering
\caption{Foundation models used in this study.}
\label{tab:fms}
\resizebox{\linewidth}{!}{%
\begin{tabular}{l l l l l}
\toprule
FM & Dimensions & Temporal Res. & Years & Source \\
\midrule
AlphaEarth & 64 & Yearly & 2020--2022 & Google Earth Engine \\
TESSERA & 128 & Yearly & 2020--2022 & GitHub  \\
\bottomrule
\end{tabular}%
        }
\end{table}

\section{Methodology}
\label{sec:methodology}

\begin{figure*}[!ht]
\centering
\includegraphics[width=0.8\linewidth]{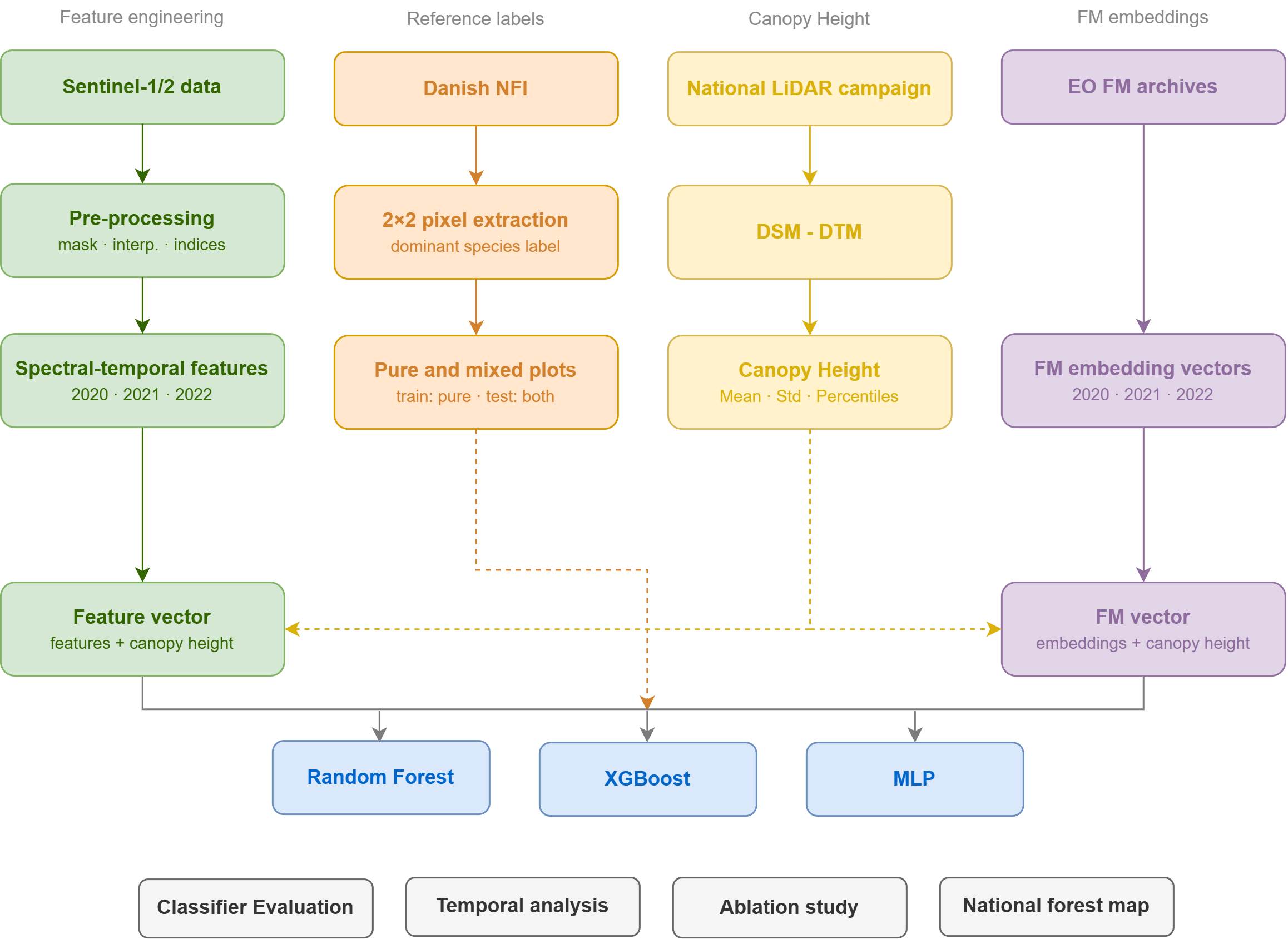}
\caption{Workflow for national-scale dominant tree species mapping. Danish NFI plots are assigned dominant species-group labels and categorized as pure or mixed. The four central $10\,\mathrm{m} \times 10\,\mathrm{m}$ pixels overlapping each plot are used to extract STF, FM embeddings, and canopy-height metrics. Models are trained using pure plots and evaluated separately on held-out pure and mixed plots after aggregating pixel-level predictions to the plot level. The best-performing model is subsequently applied across Denmark.}
\label{fig:methodology}
\end{figure*}

An overview of the methodology used in this study is illustrated in Figure~\ref{fig:methodology}.
The task is formulated as a pixel-based, multi-class classification problem, in which each pixel is assigned a tree species class label derived from the Danish NFI. 

Three ML classifiers, RF \citep{breiman_random_2001}, XGBoost \citep{Chen2016XGBoostAS}, and MLP \citep{lecun_deep_2015}, are trained on each input dataset to classify dominant forest species (Section~\ref{sec:classifiers_method}).
The results are evaluated separately for pure and mixed forest plots to provide a more ecologically grounded assessment using standard accuracy metrics (Section~\ref{sec:validation}). 
The standard accuracy assessment is complemented by an ablation study that quantifies the relative contribution of the individual input components, the influence of the acquisition year, and the sensitivity of the comparison between STF and FM embeddings to the size of the training set.
The best-performing model configuration is applied at the national scale to generate a wall-to-wall tree species map over Denmark (Section~\ref{sec:map}). 
The resulting map is accompanied by a dedicated accuracy assessment following established area estimation and map accuracy guidelines from \cite{olofsson_2014}, providing a rigorous evaluation of the operational quality of the final product.
The following subsections detail each component of the methodology, and the corresponding results are reported in Section~\ref{sec:results}.

\subsection{Input Features}
\label{sec:input_representations}
\subsubsection{Spectral-temporal features}
\label{sec:features_preprocessing_method}

Manually engineered features are constructed by concatenating Sentinel-2 interpolated spectral bands and indices, the Sentinel-2 seasonal statistics, and the Sentinel-1 monthly statistics, as described in Sections \ref{sec:sentinel_2} and \ref{sec:sentinel_1}.
We use inputs from years 2020, 2021, and 2022, as each year may hold different information for the tree species, depending on the cloud contamination, phenological status of each season, and climate. 
Incorporating observations from previous years has been shown to improve forest biomass mapping \citep{hemmerling2021mapping, zald2014influence,pflugmacher2012using}.
These features are further complemented by the static canopy height information layers, resulting in a final input vector of 3,387 features. 

\subsubsection{Foundation Model Embeddings}
\label{sec:fm_embeddings_method}

Pre-computed embeddings from AlphaEarth and TESSERA are used as the second input representation strategy evaluated in this study. 
For each pixel, embeddings are retrieved for years 2020, 2021, and 2022, consistent with the temporal range used for the manually engineered features. 
As both models produce annual embeddings, no temporal aggregation is required; instead, the embeddings from all three years are concatenated along the feature dimension to form a single multi-year representation per pixel. 
The resulting multi-year feature vectors are subsequently complemented by the static canopy height layers.
This yields final input vectors of 199 dimensions for AlphaEarth ($64 \times 3 + 7$) and 391 dimensions for TESSERA ($128 \times 3 + 7$). 
These concatenated representations are passed directly to the downstream classifiers without any fine-tuning or additional pre-processing.

\subsection{NFI tree species labeling and splits}
\label{sec:label_aggregation_method}

The NFI provides species-level labels associated with a 15 m radius circular forest plot (Section~\ref{sec:ndfi}).
To match the plot-level data to the 10 m resolution of the RS datasets, we extract the four central 10 m pixels overlapping the circular plot, corresponding to a 2×2 pixel block centered on each inventory plot, avoiding edge-contaminated pixels. We first assign each of the four pixels the same plot-level label, defined as the dominant tree species of the highest canopy layer within the block. This aggregation is justified under the assumption that the dominant species of the upper canopy layer within the plot is the primary determinant of the observed spectral and structural signal at this spatial scale.

Nonetheless, as described in  Section~\ref{sec:ndfi}, NFI plots are defined as pure or mixed based on the fractional cover of the dominant tree species in the highest canopy layer. Pure plots are more homogeneous, and therefore there is more confidence in the tree species label for the four central plot pixels. Mixed NFI plots, however, are heterogeneous, resulting in far greater uncertainty when assigning the tree species label to the four individual plot pixels. We therefore process and use pure and mixed plots in different ways for training and validation.

\subsubsection{Training plots and labels}
\label{sec:training_plots}

Due to the uncertainty when assigning pixel labels for mixed NFI plots, all classifiers are trained using pure tree species plots only, aiming to reduce the introduction of label noise in the training procedure. Training is done using the four central plot pixels with assigned tree species labels. From an ML perspective, this strategy increases the number of training samples while maintaining reliable reference labels. To avoid spatial correlation between training and validation datasets, splits are made at the plot level before extracting the four central plot pixels. Consequently, all four pixels associated with a given NFI plot are assigned to the same data split. The pure NFI plots are split randomly per class, where 80\% are allocated for training and the remaining 20\% are retained for testing (Table~\ref{tab:distribution_labels}). Within the training partition, 10\% of samples are used for model optimization and hyperparameter tuning.

\subsubsection{Test plots and labels}
\label{sec:validation_plots}

Model comparisons and final map validation are made using the test set, including: i) the remaining 20\% pure NFI plot, and ii) a sample of mixed NFI plots. After the selection of pure samples for training, the remaining NFI plots are overrepresented by mixed plots. To avoid bias towards mixed plots, we ensure that the ratio of pure and mixed plots in the test set is proportional to the pure/mixed plot ratio before training plot selection. This is done by selecting a stratified random sample of 20\% of available mixed plots per species group – matching the percentage validation split of the pure NFI plots. The inclusion of mixed plots enables evaluation of dominant species accuracy under more heterogeneous canopy conditions. The resulting class distribution is reported in  Table~\ref{tab:distribution_labels}.

\begin{table}[!ht]
\centering
\caption{Number of NFI plots used for model training and testing by tree species group. Pure plots are split into training and test sets, while mixed plots are used only for testing.}
\label{tab:distribution_labels}
\footnotesize
\begin{tabular}{l l l l l}
\toprule
Tree Species & Genus  & Training  & \multicolumn{2}{c}{Test} \\
\cmidrule(lr){4-5}
 &  &  & Pure & Mixed \\
\midrule
Beech & Fagus & 410 & 103 & 87  \\
Birch & Betula  & 145 & 36 & 57  \\
Fir & Abies & 272 & 68 & 81  \\
Larch & Larix & 56 & 14 & 32  \\
Maple & Acer & 52 & 13 & 25  \\
Oak & Quercus & 267 & 67 & 85 \\
Pine & Pinus & 245 & 61 & 67  \\
Spruce & Picea & 593 & 148 & 128 \\
OBL & -- & 106  & 27 & 58  \\
\bottomrule
\end{tabular}
\end{table}

During model evaluation and map validation, pixel-level predictions within each 2 × 2 block are aggregated to the plot level, ensuring consistency with the spatial scale at which the NFI labels are defined. A sample unit is considered correctly classified when at least two of the four central plot pixel predictions match the plot reference label (See Figure~\ref{fig:methodology}). This majority-vote aggregation is motivated by the nature of the labeling scheme: each block is assigned the label of the dominant species within the plot; however, individual pixels may not all correspond to the prevailing canopy class, as other tree species or non-tree cover may be present within the plot – especially so for mixed NFI plots. Therefore, pixel-level misclassifications may reflect label ambiguity rather than true model error. Aggregating predictions to the block level reduces the influence of this sub-plot heterogeneity, ensuring that reported classification performance reflects the ability of the model to identify the dominant forest species at the scale of the reference data.

\begin{figure*}[!ht]
\centering
\includegraphics[width=0.85\linewidth]{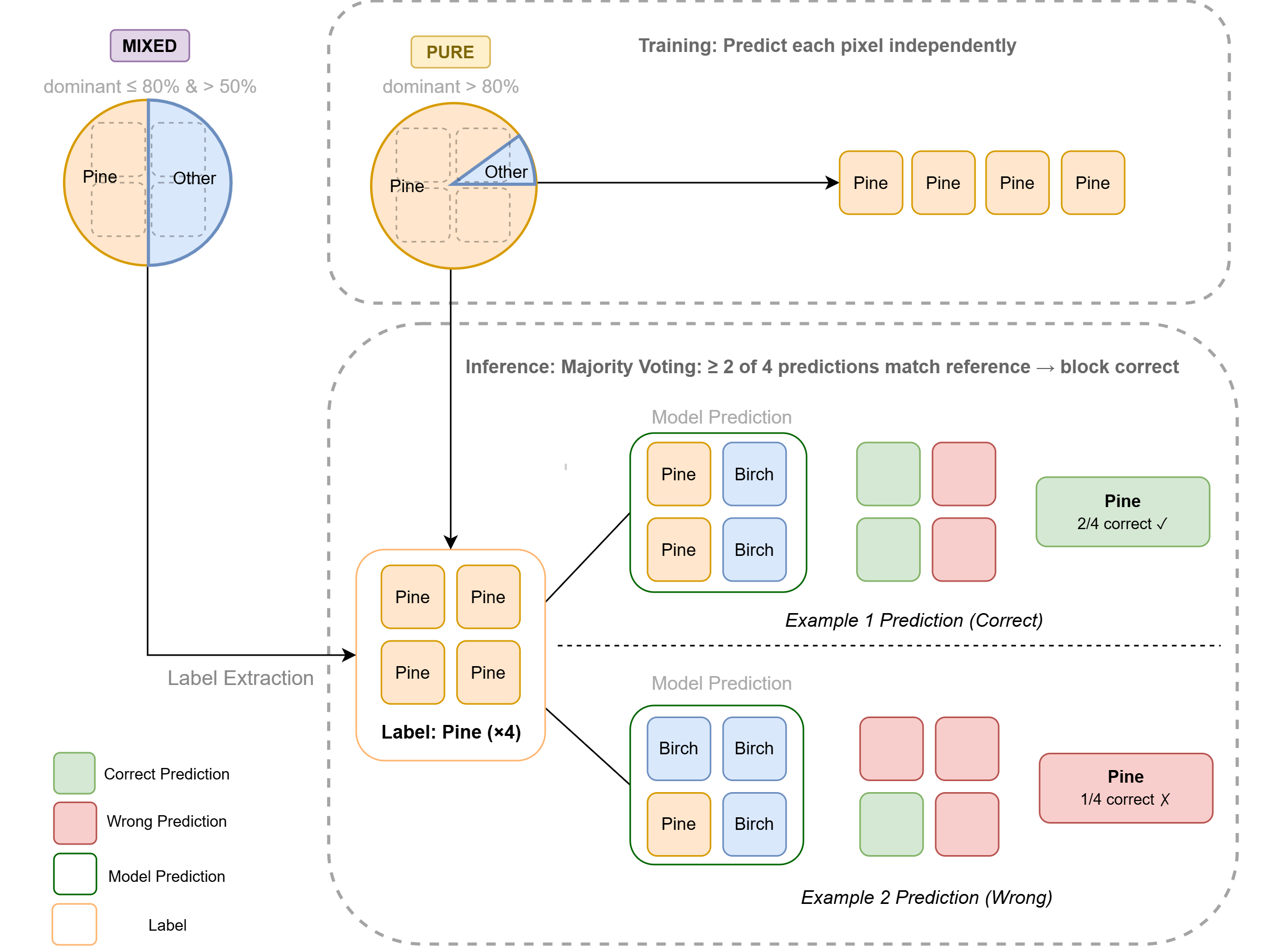}
\caption{Extraction of training and evaluation samples from NFI plots. Each plot is represented by the four overlapping $10\,\mathrm{m} \times 10\,\mathrm{m}$ pixels, which inherit the plot-level dominant tree species label. During testing, pixel-level predictions are aggregated to the plot level, and a plot is considered correctly classified when at least two of its four pixel predictions match the reference label.}\label{fig:plots}
\end{figure*}

\subsection{Classifiers}
\label{sec:classifiers_method}

Three ML classifiers are employed to map tree species from the extracted input representations: RF, XGBoost, and MLP. 
RF is an ensemble method that aggregates predictions from a large number of decision trees, offering robust performance across high-dimensional and multicollinear feature spaces \citep{BELGIU201624}. 
XGBoost is a gradient boosting framework that has demonstrated state-of-the-art performance on tabular and high-dimensional data across a broad range of supervised learning benchmarks, including RS classification tasks \citep{shwartz-ziv_tabular_2021}. 
The MLP is a feedforward neural network composed of multiple fully connected layers with nonlinear activation functions, capable of approximating complex nonlinear mappings between input features and target classes \citep{zhu_2017}. 
All MLP models consist of two hidden layers with ReLU activations \citep{Agarap2018DeepLU}, trained using the AdamW optimizer \citep{loshchilov2017decoupled} to minimize the cross-entropy loss \citep{goodfellow2016deep}, with dropout regularization \citep{JMLR:v15:srivastava14a} at a rate of 0.5 applied after each hidden layer to mitigate overfitting. 

For each classifier, key hyperparameters are tuned independently per experimental configuration using 10\% of the training samples: the number of trees and maximum features per split for RF; learning rate, maximum tree depth, and number of estimators for XGBoost; and the number of neurons per layer, learning rate, and weight decay coefficient for the MLP. 
All classifiers are evaluated under identical experimental conditions and input feature sets, ensuring a consistent and fair comparison across configurations.

\subsection{Validation}
\label{sec:validation}

Model performance comparing classifiers and different feature sets is evaluated using a set of standard accuracy metrics. For each tree species class, and for pure and mixed plots, we report Producer's Accuracy (PA), also known as recall, User's Accuracy (UA), also known as precision, and F1 score.
For overall performance assessment, macro-averaged variants of all three metrics are reported, treating each class equally regardless of its sample size. 
Overall accuracy (OA) is additionally reported as a reference metric.

The best-performing model is chosen for final map production. Following \cite{olofsson_2014}, a systematic sample may be analyzed using map classes as post-strata for estimating map accuracy and class areas, although variance estimates for the systematic design are approximate. We use the mapac R package from \cite{pflugmacher2026mapac} to report area-adjusted PA, UA, F1 score, and OA for the final map, together with per-class area estimates. All accuracy statistics and area estimates are accompanied by standard errors.

\section{Results}
\label{sec:results}

\subsection{Spectral-Temporal Features vs Foundation Model Embeddings}

The classification results for all input feature sets and classifiers are reported for pure (Table~\ref{tab:results_pure}) and mixed (Table~\ref{tab:results_mixed}) tree species plots.

\subsubsection{Pure test plots}

\begin{table*}[!ht]
\centering
\renewcommand{\arraystretch}{1.20}
\caption{Performance of all models on the pure forest test set. Per-class F1 scores are reported together with overall accuracy, macro-averaged F1 score, producer's accuracy, and user's accuracy. Bold values indicate the highest score in each row.}
\label{tab:results_pure}
\resizebox{0.8\linewidth}{!}{%
\begin{tabular}{lccccccccc}
\hline
& \multicolumn{3}{c}{STF}
& \multicolumn{3}{c}{TESSERA} 
& \multicolumn{3}{c}{AlphaEarth} \\
\cmidrule(lr){2-4}\cmidrule(lr){5-7}\cmidrule(lr){8-10}
& RF & XGBoost & MLP
& RF & XGBoost & MLP
& RF & XGBoost & MLP \\
\hline
\textbf{Tree Species} &&&&&&&&&\\
Beech  & 0.921 & 0.925 & \textbf{0.948} & 0.904 & 0.926 & 0.925 & 0.829 & 0.858 & 0.887 \\
Birch  & 0.846 & 0.853 & \textbf{0.880} & 0.714 & 0.778 & 0.806 & 0.627 & 0.700 & 0.718 \\
Fir   & 0.855 & 0.878 & \textbf{0.884} & 0.781 & 0.809 & 0.868 & 0.692 & 0.748 & 0.797 \\
Larch  & 0.720 & 0.769 & \textbf{0.815} & 0.353 & 0.800 & \textbf{0.815} & 0.421 & 0.455 & 0.696 \\
Maple  & 0.556 & 0.556 & 0.727 & 0.556 & 0.696 & \textbf{0.762} & 0.000 & 0.444 & 0.600 \\
Oak    & 0.814 & 0.830 & \textbf{0.841} & 0.760 & 0.832 & 0.814 & 0.625 & 0.667 & 0.748 \\
Pine   & 0.906 & \textbf{0.926} & 0.917 & 0.887 & 0.915 & 0.902 & 0.881 & 0.864 & 0.848 \\
Spruce & 0.898 & \textbf{0.928} & 0.894 & 0.806 & 0.867 & 0.860 & 0.812 & 0.837 & 0.875 \\
OBL    & 0.432 & 0.537 & 0.680 & 0.429 & 0.558 & \textbf{0.735} & 0.211 & 0.364 & 0.471 \\
\hline
\textbf{Total} &&&&&&&&&\\
Overall Accuracy             & 0.857 & 0.877 & \textbf{0.883} & 0.793 & 0.849 & 0.860 & 0.734 & 0.769 & 0.810 \\
F1         & 0.772 & 0.800 & \textbf{0.843} & 0.688 & 0.798 & 0.832 & 0.566 & 0.660 & 0.738 \\
Producer's acc.   & 0.751 & 0.778 & \textbf{0.830} & 0.652 & 0.772 & 0.809 & 0.563 & 0.638 & 0.715 \\
User's acc.& 0.856 & \textbf{0.868} & 0.863 & 0.824 & 0.840 & \textbf{0.868} & 0.613 & 0.722 & 0.784 \\
\hline
\end{tabular}%
}
\end{table*}

Overall, the STF MLP achieves the best classification performance across all metrics, confirming that task-specific spectral and temporal features remain highly competitive for forest species mapping when sufficient labeled data are available.
Nevertheless, the TESSERA MLP achieves competitive performance, with an F1 score 1.1 percentage points (pp) below the best-performing model, although the embeddings are derived from an FM trained on general-purpose EO data without any fine-tuning. 
Furthermore, the TESSERA MLP and the STF XGBoost achieve the highest average UA among all models.
Interestingly, the TESSERA MLP also exhibits the most balanced predictive performance across species classes, with the lowest per-class F1 score being 0.735 for the OBL class. 
In contrast, AlphaEarth embeddings prove less effective in this context, with the AlphaEarth MLP trailing the best-performing model by 10.5pp in F1 score.
Overall, across all three input representations, the MLP consistently outperforms RF and XGBoost, suggesting that the representational complexity of the feature space, whether derived from engineered features or FM embeddings, is better exploited by neural network architectures than by tree-based classifiers.

At the individual class level, the STF MLP achieves the highest F1 score for the majority of species.
Exceptions are pine and spruce, where STF XGBoost performs better, and OBL and maple, where the TESSERA MLP outperforms the rest. 
Notably, the latter two classes are among the most underrepresented classes in the training dataset (Table~\ref{tab:distribution_labels}).
This pattern suggests that embeddings pre-trained on large and diverse EO archives may partially compensate for limited class-specific training data.
This finding is further examined in  Figure~\ref{fig:mlp_tessera_report}.

\subsubsection{Mixed test plots}

\begin{table*}[!ht]
\centering
\renewcommand{\arraystretch}{1.20}
\caption{Performance of all models on the mixed forest test set. Per-class F1 scores are reported together with overall accuracy, macro-averaged F1 score, producer's accuracy, and user's accuracy.}\label{tab:results_mixed}
\resizebox{0.8\linewidth}{!}{%
\begin{tabular}{lccccccccc}
\hline
& \multicolumn{3}{c}{STF}
& \multicolumn{3}{c}{TESSERA} 
& \multicolumn{3}{c}{AlphaEarth} \\
\cmidrule(lr){2-4}\cmidrule(lr){5-7}\cmidrule(lr){8-10}
& RF & XGBoost & MLP
& RF & XGBoost & MLP
& RF & XGBoost & MLP \\
\hline
\textbf{Tree Species} &&&&&&&&&\\
Beech  & 0.680 & 0.710 & \textbf{0.767} & 0.699 & 0.711 & 0.720 & 0.635 & 0.658 & 0.676 \\
Birch  & 0.565 & \textbf{0.618} & 0.608 & 0.538 & 0.542 & 0.581 & 0.495 & 0.485 & 0.496 \\
Fir   & 0.659 & 0.666 & 0.666 & 0.588 & 0.638 & \textbf{0.679} & 0.588 & 0.617 & 0.638 \\
Larch  & 0.407 & 0.518 & \textbf{0.536} & 0.254 & 0.390 & 0.500 & 0.134 & 0.257 & 0.360 \\
Maple  & 0.130 & 0.275 & \textbf{0.539} & 0.196 & 0.376 & 0.426 & 0.016 & 0.179 & 0.177 \\
Oak    & 0.604 & 0.646 & \textbf{0.688} & 0.561 & 0.591 & 0.604 & 0.461 & 0.489 & 0.533 \\
Pine   & 0.738 & 0.733 & \textbf{0.780} & 0.679 & 0.706 & 0.725 & 0.647 & 0.659 & 0.720 \\
Spruce & 0.694 & \textbf{0.716} & 0.704 & 0.605 & 0.651 & 0.692 & 0.600 & 0.620 & 0.662 \\
OBL    & 0.342 & 0.534 & \textbf{0.590} & 0.381 & 0.452 & 0.503 & 0.346 & 0.404 & 0.495 \\
\hline
\textbf{Total} &&&&&&&&&\\

Accuracy             & 0.621 & 0.658 & \textbf{0.683} & 0.577 & 0.614 & 0.644 & 0.533 & 0.559 & 0.597 \\
F1        & 0.536 & 0.602 & \textbf{0.653} & 0.500 & 0.562 & 0.603 & 0.436 & 0.485 & 0.528 \\
Producer's  acc.  & 0.530 & 0.586 & \textbf{0.637} & 0.489 & 0.545 & 0.587 & 0.442 & 0.480 & 0.524 \\
User's acc. & 0.658 & 0.683 & \textbf{0.691} & 0.651 & 0.630 & 0.646 & 0.571 & 0.565 & 0.594 \\

\hline
\end{tabular}%
}
\end{table*}

Classification performance on mixed forest plots is substantially lower across all models and input feature sets. 
Again, the STF MLP achieves the highest performance but exhibits a drop of 19pp in overall F1 score relative to pure forest plots. 
This reduction is consistent across all species and classifiers, suggesting it reflects a fundamental challenge inherent to mixed forest mapping rather than a model-specific limitation. 

Moreover, the performance gap between TESSERA MLP and STF MLP widens in mixed forest plots, increasing to 5pp in F1 score compared to the pure forest case, suggesting that TESSERA embeddings are less effective at capturing the finer spectral and structural distinctions required to classify compositionally heterogeneous stands.

As AlphaEarth consistently underperforms relative to the other feature representations, the remainder of the analysis focuses on the STF and TESSERA feature sets.

\subsubsection{Confusion Matrices}

\begin{table}[!htbp]
\centering
\caption{Row-normalized confusion matrices (\%) for the STF MLP and TESSERA MLP evaluated on the pure and mixed forest test sets. Rows correspond to reference species and columns to predicted species. Bold values indicate the highest score in each row.}
\label{tab:confmats-all}

\begin{subtable}[t]{0.49\textwidth}
\centering
\caption{STF MLP (pure)}
\label{tab:confmat-mlp-pure}
\resizebox{\textwidth}{!}{%
\begin{tabular}{lccccccccc}
\hline
True $\backslash$ Pred & Beech & Birch & Fir & Larch & Maple & Oak & Pine & Spruce & OBL \\
\hline
Beech  & \textbf{97.1} & 0.0 & 0.0 & 0.0 & 0.0 & 1.9 & 0.0 & 1.0 & 0.0 \\
Birch  & 2.8 & \textbf{91.7} & 0.0 & 0.0 & 0.0 & 2.8 & 0.0 & 0.0 & 2.8 \\
Fir    & 0.0 & 0.0 & \textbf{89.7} & 0.0 & 1.5 & 0.0 & 1.5 & 7.4 & 0.0 \\
Larch  & 0.0 & 0.0 & 7.1 & \textbf{78.5} & 0.0 & 0.0 & 0.0 & 14.3 & 0.0 \\
Maple  & 7.7 & 0.0 & 0.0 & 0.0 & \textbf{61.5} & 23.1 & 0.0 & 0.0 & 7.7 \\
Oak    & 6.0 & 3.0 & 0.0 & 1.5 & 0.0 & \textbf{86.6} & 0.0 & 1.5 & 1.5 \\
Pine   & 0.0 & 1.6 & 1.6 & 0.0 & 0.0 & 0.0 & \textbf{90.2} & 6.6 & 0.0 \\
Spruce & 0.7 & 1.4 & 4.1 & 0.7 & 0.0 & 1.4 & 1.4 & \textbf{88.5 }& 2.0 \\
OBL    & 3.7 & 3.7 & 3.7 & 0.0 & 0.0 & 18.5 & 3.7 & 3.7 & \textbf{63.0} \\
\hline
\end{tabular}%
}
\end{subtable}
\hfill
\begin{subtable}[t]{0.49\textwidth}
\centering
\caption{STF MLP (mixed)}
\label{tab:confmat-mlp-mixed}
\resizebox{\textwidth}{!}{%
\begin{tabular}{lccccccccc}
\hline
True $\backslash$ Pred & Beech & Birch & Fir & Larch & Maple & Oak & Pine & Spruce & OBL \\
\hline
Beech  & \textbf{81.4} & 2.1 & 2.3 & 0.5 & 0.7 & 5.1 & 0.5 & 5.1 & 2.5 \\
Birch  & 4.9 & \textbf{56.2} & 6.6 & 1.4 & 0.0 & 10.1 & 4.5 & 8.0 & 8.3 \\
Fir    & 3.2 & 0.0 & \textbf{73.4} & 0.5 & 0.2 & 1.5 & 2.5 & 17.9 & 0.7 \\
Larch  & 7.5 & 6.2 & 8.8 & \textbf{43.8} & 0.6 & 11.2 & 5.0 & 13.8 & 3.1 \\
Maple  & 27.8 & 0.8 & 2.4 & 0.0 & \textbf{41.3} & 13.5 & 0.8 & 4.0 & 9.5 \\
Oak    & 5.9 & 3.3 & 3.8 & 0.7 & 0.7 & \textbf{71.6} & 2.4 & 5.7 & 5.9 \\
Pine   & 0.6 & 3.0 & 5.7 & 0.6 & 0.3 & 3.0 & \textbf{77.6} & 7.2 & 2.1 \\
Spruce & 1.7 & 2.2 & 12.7 & 2.8 & 0.0 & 3.0 & 4.1 & \textbf{71.2} & 2.3 \\
OBL    & 7.6 & 8.7 & 9.7 & 0.0 & 2.1 & 11.8 & 0.7 & 2.8 & \textbf{56.6} \\
\hline
\end{tabular}%
}
\end{subtable}

\vspace{0.7em}

\begin{subtable}[t]{0.49\textwidth}
\centering
\caption{TESSERA MLP (pure)}
\label{tab:confmat-tessera-pure}
\resizebox{\textwidth}{!}{%
\begin{tabular}{lccccccccc}
\hline
True $\backslash$ Pred & Beech & Birch & Fir & Larch & Maple & Oak & Pine & Spruce & OBL \\
\hline
Beech  & \textbf{97.1} & 0.0 & 0.0 & 0.0 & 0.0 & 1.9 & 0.0 & 1.0 & 0.0 \\
Birch  & 2.8 & \textbf{75.0} & 2.8 & 0.0 & 0.0 & 2.8 & 0.0 & 5.6 & 11.1 \\
Fir    & 0.0 & 0.0 & \textbf{80.9} & 0.0 & 0.0 & 0.0 & 0.0 & 17.6 & 1.5 \\
Larch  & 7.1 & 0.0 & 0.0 & \textbf{71.4} & 0.0 & 0.0 & 0.0 & 21.4 & 0.0 \\
Maple  & 7.7 & 0.0 & 0.0 & 0.0 & \textbf{53.8} & 15.4 & 0.0 & 7.7 & 15.4 \\
Oak    & 9.0 & 1.5 & 1.5 & 1.5 & 0.0 & \textbf{83.6} & 0.0 & 3.0 & 0.0 \\
Pine   & 0.0 & 1.6 & 1.6 & 1.6 & 0.0 & 0.0 & \textbf{88.5} & 6.6 & 0.0 \\
Spruce & 2.0 & 0.0 & 4.7 & 0.0 & 0.0 & 4.7 & 2.0 & \textbf{85.8} & 0.7 \\
OBL    & 7.4 & 3.7 & 0.0 & 3.7 & 0.0 & 14.8 & 3.7 & 0.0 & \textbf{66.7} \\
\hline
\end{tabular}%
}
\end{subtable}
\hfill
\begin{subtable}[t]{0.49\textwidth}
\centering
\caption{TESSERA MLP (mixed)}
\label{tab:confmat-tessera-mixed}
\resizebox{\textwidth}{!}{%
\begin{tabular}{lccccccccc}
\hline
True $\backslash$ Pred & Beech & Birch & Fir & Larch & Maple & Oak & Pine & Spruce & OBL \\
\hline
Beech  & \textbf{75.6} & 1.4 & 2.8 & 0.2 & 0.5 & 9.2 & 0.2 & 6.2 & 3.9 \\
Birch  & 3.5 & \textbf{48.6} & 3.8 & 2.1 & 0.3 & 10.4 & 8.0 & 13.9 & 9.4 \\
Fir    & 2.7 & 0.2 & \textbf{71.0} & 0.5 & 0.0 & 3.7 & 1.7 & 19.4 & 0.7 \\
Larch  & 9.4 & 3.1 & 3.1 & \textbf{38.1} & 0.6 & 18.1 & 0.6 & 21.9 & 5.0 \\
Maple  & 28.6 & 1.6 & 4.0 & 0.0 & \textbf{25.4} & 15.1 & 0.0 & 8.7 & 16.7 \\
Oak    & 6.1 & 4.0 & 6.9 & 0.5 & 0.2 & \textbf{66.7} & 2.8 & 7.8 & 5.0 \\
Pine   & 0.9 & 1.8 & 2.7 & 1.5 & 0.0 & 5.1 & \textbf{68.1} & 17.9 & 2.1 \\
Spruce & 3.0 & 0.8 & 10.8 & 2.0 & 0.0 & 2.8 & 3.1 & \textbf{75.6} & 1.9 \\
OBL    & 7.3 & 9.4 & 6.2 & 0.3 & 1.4 & 19.4 & 1.4 & 6.9 & \textbf{47.6} \\
\hline
\end{tabular}%
}
\end{subtable}

\end{table}

The confusion matrices for the STF MLP and the TESSERA MLP are presented in Table~\ref{tab:confmats-all} for both pure and mixed forest plots. 
Despite relying on fundamentally different input feature sets, one derived from manually engineered spectral and temporal features, the other from task-agnostic FM embeddings, both models exhibit largely consistent inter-class confusion patterns across both forest types.
This structural similarity suggests that TESSERA embeddings encode discriminatory information closely aligned with that captured by handcrafted features, providing evidence of the representational richness of FM-derived inputs for tree species discrimination.

For pure forest plots, both models correctly classify the majority of species with high confidence, as reflected by the strong diagonal dominance in Tables~\ref{tab:confmat-mlp-pure} and~\ref{tab:confmat-tessera-pure}. 
The main off-diagonal patterns are broadly similar across models, although some apparently large percentages correspond to only a few plots for rare classes such as maple and larch. 
Broadleaved errors mainly involve confusion among maple, oak, and OBL, consistent with overlapping phenological trajectories and canopy reflectance properties among deciduous species. 

In mixed forest plots, classification confidence decreases substantially for both models, as evidenced by the more diffuse confusion matrices in Tables~\ref{tab:confmat-mlp-mixed} and~\ref{tab:confmat-tessera-mixed}. 
Inter-class confusions become more pronounced, with maple frequently misclassified as beech and spruce confused with fir.

\subsubsection{Limited Training Data}

\begin{figure}[!ht]
\centering
\includegraphics[width=\columnwidth]{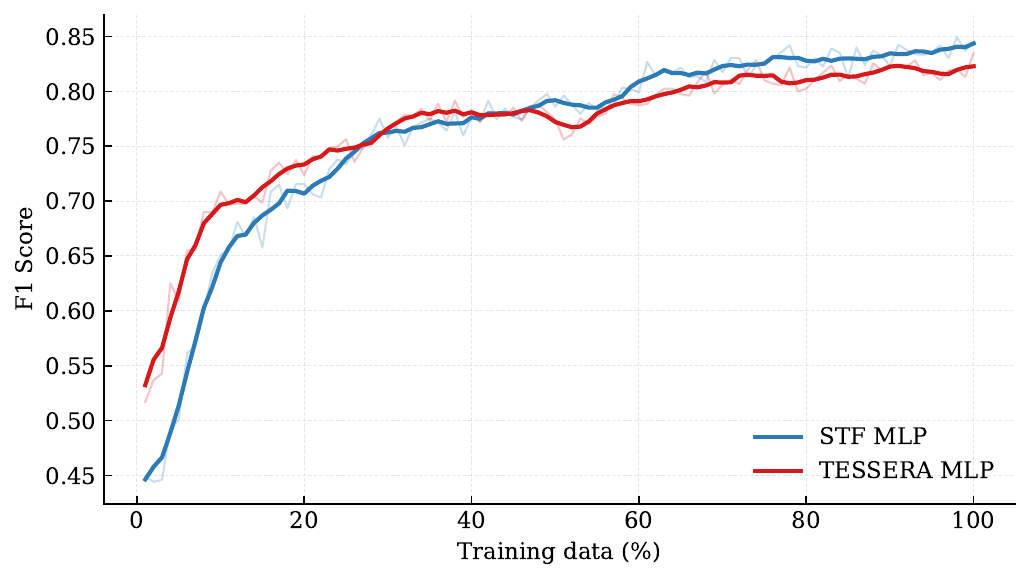}
\caption{Macro F1 scores of the TESSERA MLP and the STF MLP as a function of the percentage of available training plots used for model fitting.}
\label{fig:mlp_tessera_report}
\end{figure}

Since obtaining sufficient labeled data is often challenging in operational settings, we further examine how the two representations behave when labels are scarce.
To systematically assess this, both the TESSERA MLP and the STF MLP are trained across a range of training set sizes, from 1\% to 100\% of the available training data, and evaluated on the same held-out pure forest test set. 
The resulting F1 scores for each of the training sizes are presented in Figure~\ref{fig:mlp_tessera_report}. 
As the training set is progressively reduced, the performance of the STF MLP declines more steeply than that of the TESSERA MLP.
Below approximately 25\% of the original training data (corresponding to roughly 500 plots), TESSERA MLP consistently outperforms STF MLP, with the performance gap widening further as training data becomes increasingly scarce.

\subsection{Single year vs Multi-year input}

To assess the impact of temporal coverage on classification performance, we train the two best-performing models, the STF MLP and the TESSERA MLP, using single-year inputs, consecutive two-year combinations, and the full three-year configuration used in the main experiments. F1 scores for pure and mixed forest plots are reported in Table~\ref{tab:year_assessment}.
For both models and test sets, the three-year configuration achieves the highest classification performance. 

For the STF MLP, single-year performance varies considerably across individual-year configurations. On pure plots, the 2020 input yields the highest single-year F1 score, whereas the 2022 input yields the lowest, corresponding to a difference of 7.6\,pp. Similarly, the 2020--2021 configuration reaches a higher score compared to the 2021--2022 configuration with a difference of  4.4\,pp.
The TESSERA MLP exhibits less variation among the individual-year configurations, with F1 scores ranging from 0.771 to 0.797 on pure plots. Adding a second year results in modest improvements, whereas the full three-year configuration increases the F1 score to 0.832. TESSERA outperforms STF for the 2022 and 2021--2022 configurations, while STF performs better for 2020, 2021, and 2020--2021.
A similar pattern is observed for mixed plots. The STF MLP exhibits greater variability across the individual-year configurations, with F1 scores ranging from 0.578 to 0.635, whereas the corresponding TESSERA scores range from 0.563 to 0.583. 

These findings suggest that the year is important and that FMs are more robust to year selection compared to STF.

\begin{table}[t]
\centering
\caption{F1 scores of the STF MLP and the TESSERA MLP under different temporal input configurations. Results are shown separately for pure and mixed forest test plots.}
\label{tab:year_assessment}
\begin{tabular}{lcccc}
\toprule
\textbf{Years} & \multicolumn{2}{c}{\textbf{STF MLP}} & \multicolumn{2}{c}{\textbf{TESSERA MLP}}\\
\cmidrule(lr){2-3}\cmidrule(lr){4-5}
 & Pure & Mixed & Pure & Mixed \\
\midrule
2020         & 0.819 & 0.635 & 0.797 & 0.577 \\
2021         & 0.802 & 0.615 & 0.771 & 0.563 \\
2022         & 0.743 & 0.578 & 0.793 & 0.583 \\
2020, 2021   & 0.831 & 0.651 & 0.798 & 0.593 \\
2021, 2022   & 0.787 & 0.618 & 0.809 & 0.586 \\
2020, 2021, 2022   & 0.843 &  0.653 & 0.832 & 0.603 \\
\bottomrule
\end{tabular}
\end{table}

\subsection{Ablation Study}

The STF representation combines multi-temporal Sentinel-2 spectral bands with three supplementary feature groups: Sentinel-1 SAR backscatter, canopy height metrics, and spectral indices. To quantify their contribution to classification performance, we conduct an ablation study using the best-performing STF MLP model.
Specifically, we systematically remove each input source from the full feature set and retrain the model. 
Results are reported in Table~\ref{tab:ablation_multicol_checks_rev} separately for pure and mixed forest plots.

\begin{table}[!ht]
\centering
\caption{Ablation study for the STF MLP showing the effect of Sentinel-1 (S1), canopy height (CH), and Sentinel-2 spectral indices on performance. F1 scores are reported separately for the pure and mixed forest test sets.}

\label{tab:ablation_multicol_checks_rev}
\small
\setlength{\tabcolsep}{10pt}
\begin{tabular}{cccccc}
\hline
\textbf{S1} & \textbf{CH} & \textbf{Indices} & \multicolumn{2}{c}{\textbf{F1}} \\
\cmidrule(lr){4-5}
 &  &  & \textbf{Pure} & \textbf{Mixed} \\
\hline
\checkmark & \checkmark & \checkmark & 0.843 & 0.653 \\
\checkmark & \checkmark & -- & 0.818 & 0.631 \\
\checkmark & -- & \checkmark & 0.830 & 0.644 \\
\checkmark & -- & -- & 0.823 & 0.637 \\
-- & \checkmark & \checkmark & 0.793 & 0.619 \\
-- & \checkmark & -- & 0.790 & 0.617 \\
-- & -- & \checkmark & 0.793 & 0.625 \\
-- & -- & -- & 0.790 & 0.617 \\
\hline
\end{tabular}
\end{table}

The full feature set, combining all four components, achieves the highest performance for both pure and mixed tree cover, with F1-Scores of 0.843 and 0.653 for pure and mixed plots, respectively, confirming that all input sources are important for achieving the highest classification performance.
Among the individual components, S1 data emerge as the most influential, providing the largest single-source improvement over the Sentinel-2 baseline for both forest types, an increase of 5.7pp for pure plots and 2.0pp for mixed plots. 

This highlights the important role of SAR-derived information in forest species discrimination, which we attribute to its sensitivity to canopy structure and its ability to provide cloud-independent observations during spectrally critical phenological periods that may be obscured in optical imagery.
Canopy height and spectral indices contribute more modestly but consistently, with their removal resulting in incremental performance reductions across both forest categories. 
Moreover, the combined removal of all three supplementary components, retaining only S2 spectral bands, results in a pronounced degradation of 7.2pp for pure plots and 3.5pp for the mixed plots, suggesting that the additional input sources are particularly beneficial for forest species mapping.

\subsection{Forest Maps}
\label{sec:map}

\subsubsection{Visual evaluation}

\begin{figure*}[!ht]
    \centering
    \caption{Representative examples of the final tree species map in two forest areas in Denmark. (a) shows a spring 2022 orthophoto, (b) shows available stand-level tree-species polygons from the Danish Nature Agency used for qualitative comparison, and (c) shows the corresponding prediction from the final model.}
    \includegraphics[width=1\linewidth]{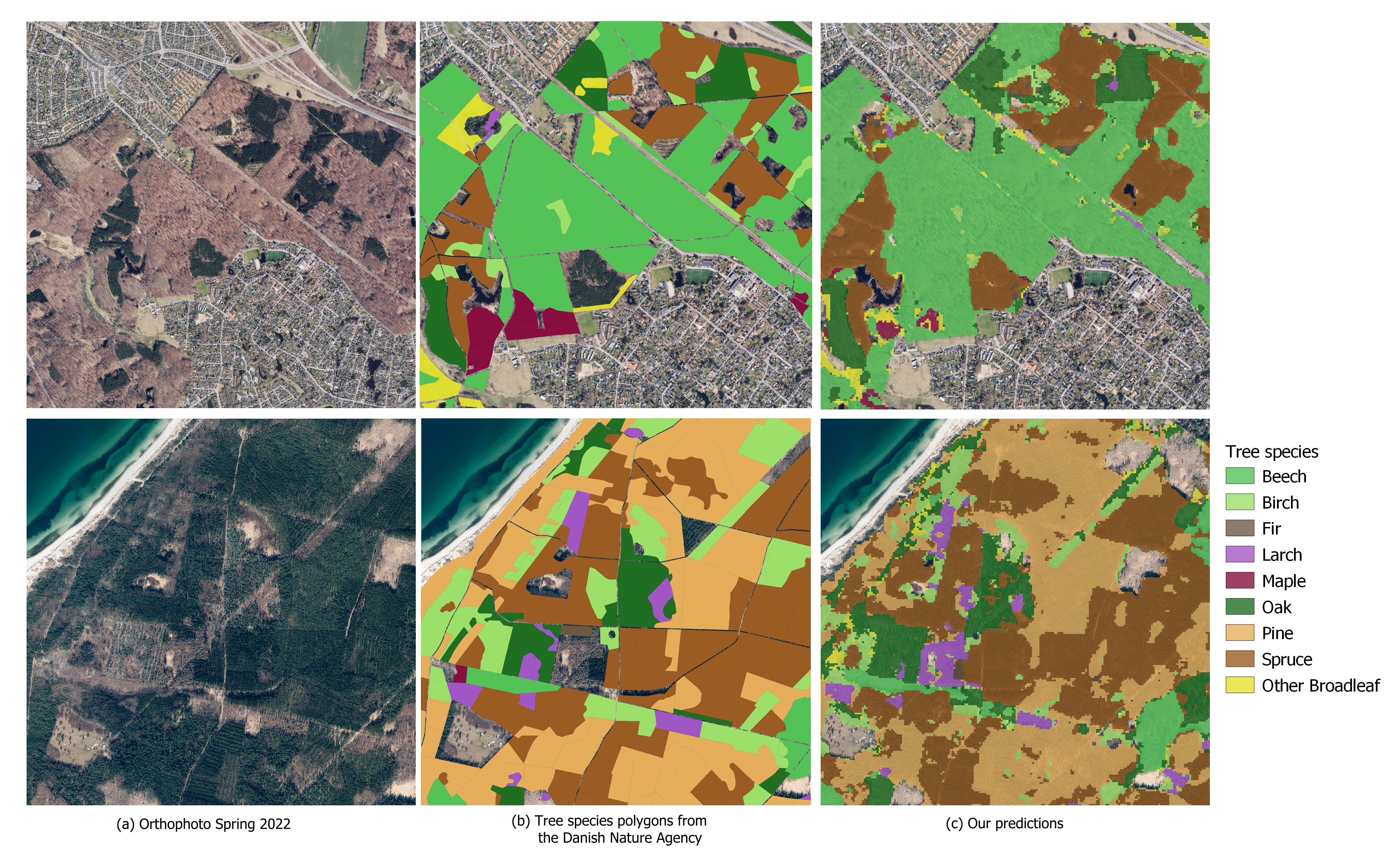}
    \label{fig:example_map}
\end{figure*}

One of the primary contributions of this study is the publication of the first national-scale tree species map of Denmark at 10\,m spatial resolution. 
The map is generated by applying the best-performing model, STF MLP, trained on Sentinel-1 and Sentinel-2 time series across the entire country. 

Figure~\ref{fig:example_map} presents two examples of the final tree species map in forest landscapes, alongside complementary reference layers. Stand-level tree species polygons from the NST are included only to provide qualitative visual context, and are not used as an independent quantitative validation dataset, as their spatial support and thematic definition differ from those of the NFI-based reference labels and the pixel-based predictions.

Overall, the predicted map reproduces the main spatial pattern of forest stands visible in the orthophotos and broadly agrees with the dominant species patterns represented in the NST polygons. 
Importantly, pixel-level predictions capture finer within-stand spatial variation than the polygon-based layer, particularly along stand boundaries and in structurally heterogeneous forest areas.
In addition, the wall-to-wall nature of the predictions provides spatially continuous tree species information in areas where the NST polygons are absent, highlighting the practical value of the map for comprehensive national forest monitoring beyond the spatial coverage of existing polygon datasets.

\subsubsection{Mapped area comparison}

\begin{figure*}[!htbp]
    \centering
    \includegraphics[width=\linewidth]{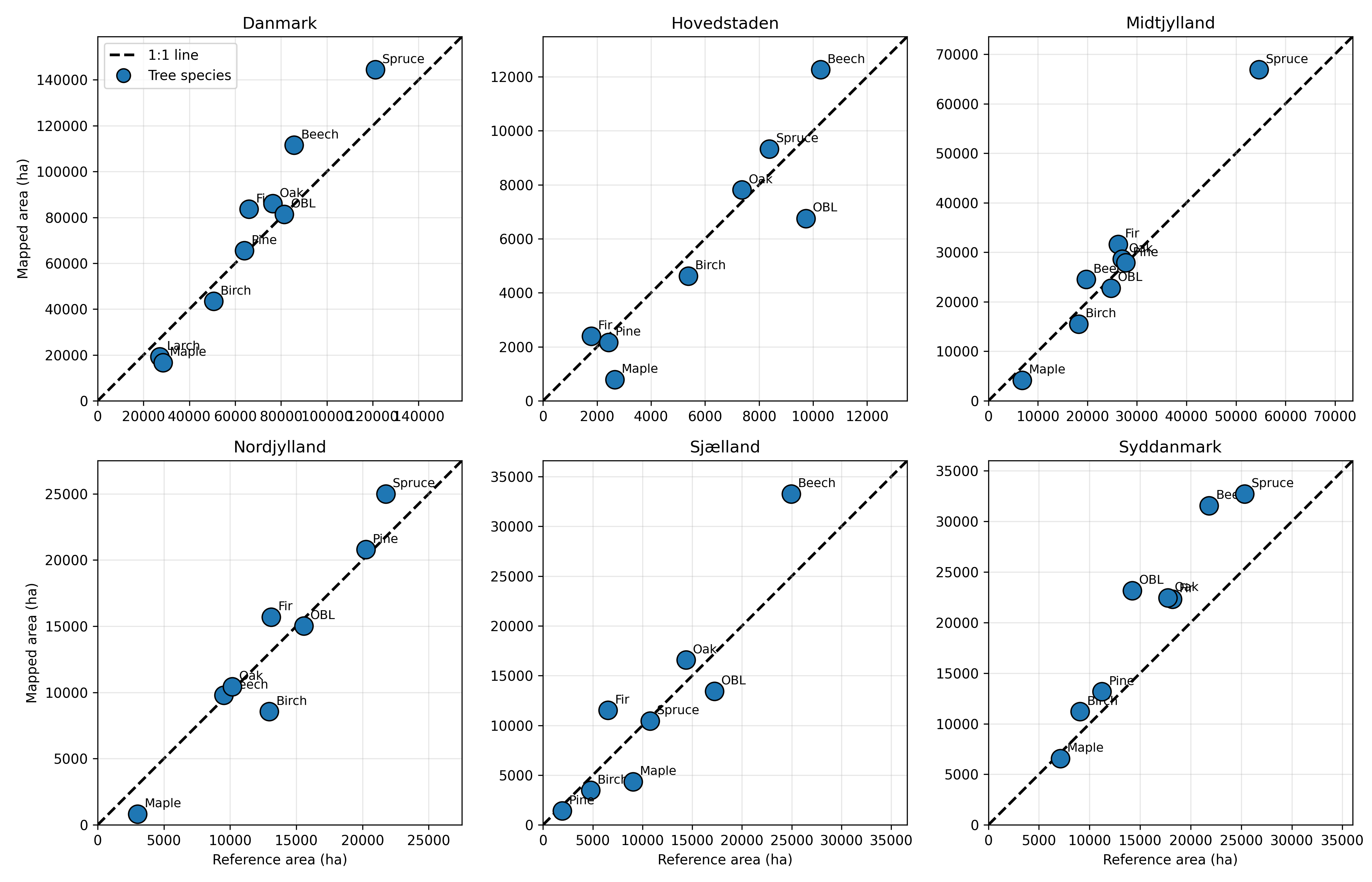}
    \caption{Mapped area of tree species at the national and regional level. Areas are derived from the final map (y-axis) and compared with the corresponding Danish National Forest Statistics for 2022 reported by \cite{Skovstat2022}. For larch, no corresponding statistics are available in the 2022 report; therefore, only the national estimate from the 2023 report \citep{NordLarsenEtAl2025} is included, and no regional comparison is provided.}
    \label{fig:area_stats}
\end{figure*}

Figure~\ref{fig:area_stats} summarizes the mapped area in hectares (ha) of tree species and the corresponding areas reported in the Danish National Forest Statistics for 2022 \citep{Skovstat2022}, both at the national level and for each of the five main regions of Denmark. To ensure comparability, several categories from the national statistics are aggregated to match the classes used in this study. Specifically, spruce comprises \textit{Rødgran} and \textit{Sitkagran}, fir comprises \textit{Andet ædelgran}, \textit{Nordmannsgran}, and \textit{Nobilis}, and OBL comprises \textit{Ask} and \textit{Andet løv}. The category \textit{Other conifers} (41,681 ha nationally) is excluded from the NFI reference because it could not be assigned consistently to any of the mapped classes. For larch, no comparable regional statistics are available; therefore, only the national estimate reported in \cite{NordLarsenEtAl2025} is included for comparison.

The comparison provides an indication of how closely the mapped species distribution aligns with the official statistics. 
Overall, pine and OBL agree well with the national statistics (within $\pm3\%$), while oak and birch deviate moderately in relative terms ($+13\%$ and $-14\%$) but by less than 10,000~ha in absolute area. The largest positive discrepancies occur for beech ($+25{,}900$~ha), spruce ($+23{,}300$~ha), and fir ($+17{,}600$~ha), which together dominate the overall overestimation. In contrast, maple and larch are slightly underestimated ($-11{,}700$ and $-7{,}800$~ha); although their relative deviations are large, their small national extent limits their absolute contribution.
For the conifer classes, part of the overestimation may be explained by the exclusion of the \textit{Other conifers} category from the mapping framework. Although this category accounts for 41,681 ha in the national statistics, no corresponding map class was available, requiring these stands to be assigned to one of the mapped species classes and potentially increasing the estimated areas of other conifer species. The total forest area estimated from the map is 651,855 ha, compared with 642,380 ha in the reference statistics, including the \textit{Other conifers} category. This relatively small difference suggests that the dominant-species mapping approach at 10\,m spatial resolution produces area estimates that are broadly consistent with inventory-based estimates.

At the regional level, the closest correspondence is observed in Nordjylland, with good agreement also in Midtjylland (the largest region), where many species fall within or close to the $\pm10\%$ range despite an overestimation of spruce. In contrast, Syddanmark exhibits the largest discrepancies, with most species showing deviations greater than 10\% from the reference statistics. Intermediate levels of agreement are observed for Hovedstaden and Sjælland, where some species align well with the reference data while others are substantially overestimated.

\subsubsection{Map accuracy assessment}

The area-adjusted accuracy assessment (Table~\ref{tab:area_adjusted}) provides unbiased estimates of map classification accuracy and standard errors. 
For this, we combined both the pure and the mixed test plots to perform a representative evaluation of the produced map. 
The overall accuracy of the map is estimated at 79.9\% ($\pm1.3\%$), indicating a generally robust classification performance at the map level. 

\begin{table*}[htbp]
\centering
\caption{Area-adjusted confusion matrix, accuracy estimates and standard errors for the final national tree species map.}
\label{tab:area_adjusted}
\scriptsize
\setlength{\tabcolsep}{2pt}
\begin{tabular}{l
                S[table-format=2.1] S[table-format=2.1] S[table-format=2.1]
                S[table-format=2.1] S[table-format=2.1] S[table-format=2.1]
                S[table-format=2.1] S[table-format=2.1] S[table-format=2.1]
                S[table-format=2.1] S[table-format=2.1]
                l l l}
\toprule
 & \multicolumn{9}{c}{\textbf{Reference class (area \%)}} &
 \multicolumn{2}{c}{$\Sigma$} &
 \multicolumn{3}{c}{\textbf{Accuracy \%}} \\
\cmidrule(lr){2-10} \cmidrule(lr){11-12} \cmidrule(lr){13-15}

\textbf{Species}
 & {Beech} & {Birch} & {Fir} & {Larch} & {Maple} & {Oak} & {Pine} & {Spruce} & {OBL}
& {Map} & {Ref.} &
\multicolumn{1}{c}{PA} &
\multicolumn{1}{c}{UA} &
\multicolumn{1}{c}{F1} \\
\midrule

Beech
& \textbf{14.7} & 0.3 & {--} & 0.2 & 0.6 & 0.6 & {--} & 0.3 & 0.5 
& 17.1 & 15.9
& 92.4 $\pm$ 2.1 & 85.8 $\pm$ 2.4 & 89.0 $\pm$ 1.6 \\

Birch
& 0.2 & \textbf{5.2} & {--} & {--} & {--} & 0.2 & 0.3 & 0.3 & 0.4 
& 6.7 & 7.4
& 70.2 $\pm$ 4.7 & 78.0 $\pm$ 4.6 & 73.9 $\pm$ 3.3 \\

Fir
& {--} & 0.1 & \textbf{10.6} & 0.2 & {--} & 0.1 & 0.2 & 1.6 & {--} 
& 12.8 & 13.0
& 81.4 $\pm$ 2.9 & 82.5 $\pm$ 3.4 & 81.9 $\pm$ 2.2 \\

Larch
& {--} & 0.1 & 0.1 & \textbf{2.3} & {--} & 0.1 & {--} & 0.3 & {--} 
& 3.0 & 3.8
& 62.4 $\pm$ 6.3 & 79.3 $\pm$ 7.5 & 69.8 $\pm$ 4.9 \\

Maple
& {--} & {--} & {--} & 0.1 & \textbf{2.2} & {--} & {--} & {--} & 0.3 
& 2.6 & 4.2
& 51.9 $\pm$ 6.7 & 85.0 $\pm$ 8.0 & 64.4 $\pm$ 5.7 \\

Oak
& 0.4 & 0.6 & 0.1 & 0.3 & 0.5 & \textbf{9.8} & 0.1 & 0.2 & 1.1 
& 13.2 & 12.7
& 77.0 $\pm$ 3.8 & 74.3 $\pm$ 3.5 & 75.7 $\pm$ 2.6 \\

Pine
& {--} & 0.3 & 0.2 & 0.1 & {--} & 0.2 & \textbf{8.6} & 0.6 & 0.2 
& 10.0 & 9.8
& 87.5 $\pm$ 3.2 & 85.5 $\pm$ 3.2 & 86.5 $\pm$ 2.3 \\

Spruce
& 0.4 & 0.4 & 1.9 & 0.5 & 0.1 & 0.5 & 0.4 & \textbf{17.8} & 0.2 
& 22.2 & 21.7
& 81.8 $\pm$ 2.4 & 80.2 $\pm$ 2.4 & 81.0 $\pm$ 1.7 \\

OBL
& 0.2 & 0.4 & 0.2 & {--} & 0.8 & 1.2 & 0.2 & 0.6 & \textbf{8.7} 
& 12.5 & 11.4
& 76.4 $\pm$ 3.5 & 70.0 $\pm$ 5.9 & 73.0 $\pm$ 3.6 \\

\midrule
\multicolumn{15}{r}{\textbf{Overall accuracy} = 79.9 $\pm$ 1.3} \\
\bottomrule
\end{tabular}
\end{table*}

Accuracy metrics vary across species classes and are broadly consistent with the patterns observed in the previous classification experiments.
Beech shows the highest agreement between mapped and reference data (PA: $92.4\%$, UA: $85.8\%$), confirming its strong and stable separability already evident from the high per-class F1 scores across models. 
Similarly, pine, fir, and spruce achieve consistently high accuracies, in line with their robust performance in both pure and mixed settings, suggesting that these dominant coniferous classes are well captured. 

In contrast, lower PA is observed for maple and larch, indicating higher omission errors. This pattern is also reflected in Figure~\ref{fig:area_stats}, where both species show substantial underestimation relative to the national forest statistics.
This is consistent with their comparatively lower F1 scores and higher confusion levels in the classification results, particularly for underrepresented classes and in mixed stands.
Despite this, their relatively high user accuracies suggest that when these classes are predicted, they are generally reliable, reflecting the conservative behavior also observed in the model outputs. 

\begin{figure}[!ht]
\centering
\includegraphics[scale=0.5]{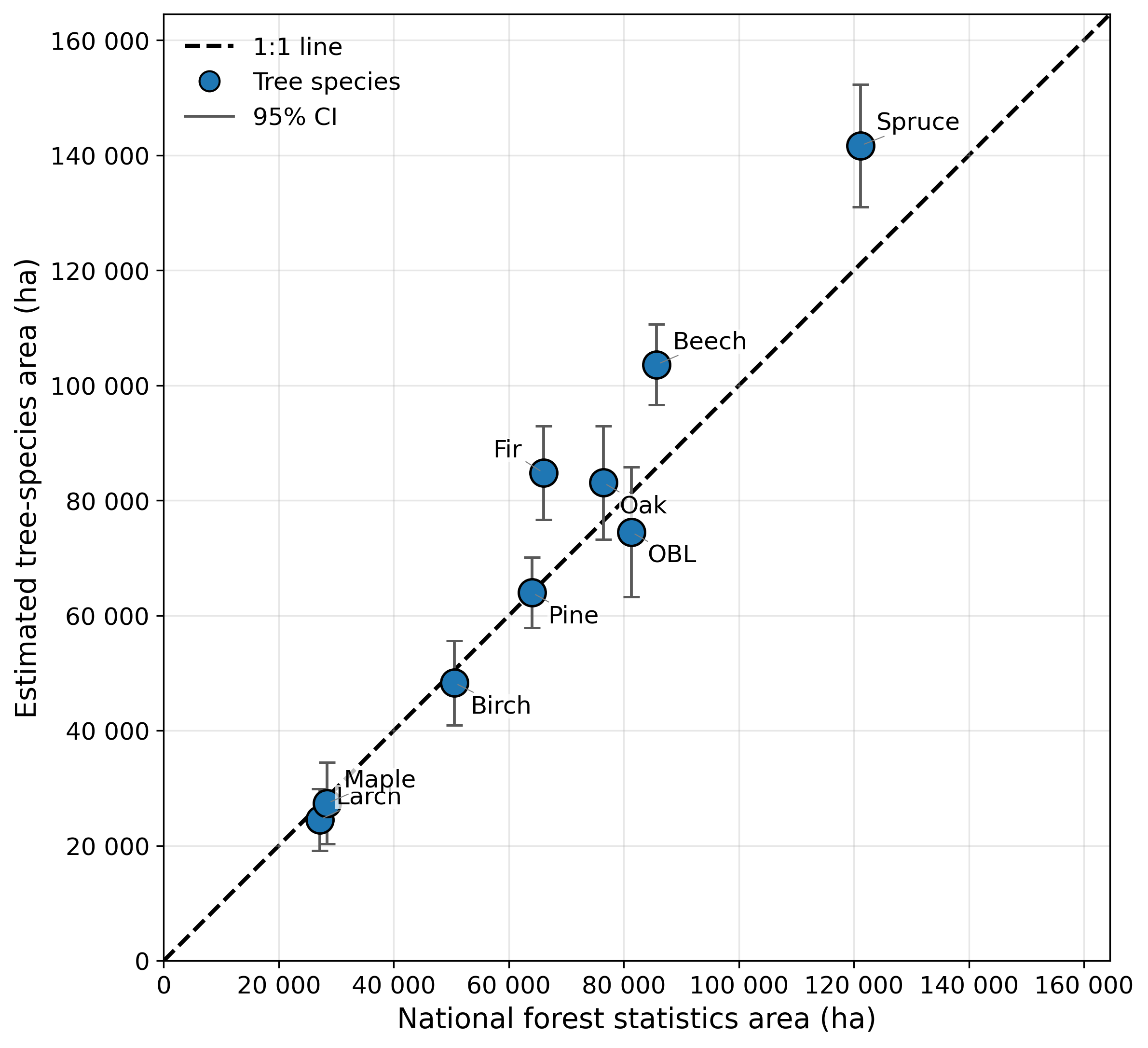}
\caption{Area estimates of tree species using the validation sample and mapped classes as strata \citep{olofsson_2014} compared with the corresponding NFI statistics for 2022 reported by \cite{Skovstat2022}. For larch, no corresponding statistics are available in the 2022 report; therefore, only the national estimate from the 2023 report \citep{NordLarsenEtAl2025} is included, and no regional comparison is provided.}
\label{fig:areas_compare_olofson}
\end{figure}

Figure~\ref{fig:areas_compare_olofson} compares the area-adjusted estimates derived using the approach of \cite{olofsson_2014} with the Danish national forest statistics. Compared with the direct map-based estimates shown in Figure~\ref{fig:area_stats}, the adjusted areas exhibit closer agreement with the reference statistics. Most species fall within a $\pm10\%$ agreement range, indicating that accounting for classification errors through area-adjusted estimation largely removes the biases observed in the map-derived areas. While spruce, fir, and beech remain overestimated, the area-adjusted estimates move considerably closer to the reference values. For beech, the discrepancy is reduced by approximately 8,000~ha compared with the direct map-based estimate. The 95\% confidence intervals overlap the reference statistics for most species, further suggesting that the remaining deviations are small relative to the uncertainty of the area estimates.
The reduced discrepancies observed after area adjustment are consistent with the omission and commission errors quantified in Table~\ref{tab:area_adjusted}, indicating that a substantial fraction of the differences observed in Figure~\ref{fig:area_stats} can be attributed to classification error rather than systematic bias in the mapped distribution of tree species.

\section{Discussion}
\label{sec:discussion}

\subsection{Comparison of spectral-temporal features with Foundation Models}

\subsubsection{Classification Performance of STF and FM Approaches} 

These results indicate that manually engineered spectral-temporal representations remain particularly effective for fine-grained tree species discrimination, especially under compositionally heterogeneous forest conditions. 
The advantage of STF likely reflects its explicit preservation of dense phenological information from Sentinel-2, together with complementary structural information from Sentinel-1 and canopy-height data. 
In contrast, the annual TESSERA embeddings provide a more temporally aggregated representation and may therefore retain less of the fine-scale seasonal variation needed to distinguish species in mixed stands. 

At the species level, both representations exhibit broadly similar confusion patterns, suggesting that several errors arise from limitations shared across the input representations rather than from one model alone.
The larch--spruce confusion should be interpreted cautiously because the pure test set contains only 14 larch plots, meaning that a small number of errors results in large changes in the corresponding percentages. 
The tendency of the TESSERA MLP to confuse fir with spruce is more directly consistent with the well-documented difficulty of separating closely related coniferous genera using RS observations \citep{fassnacht2016review, immitzer2012tree}.
Maple and beech often co-exist within mixed broadleaved stands, which can explain the amplification of their confusion in the mixed plots.
The consistency of confusion patterns across both models and forest types suggests that misclassifications are driven largely by species co-occurrence, spectral or structural similarity, and the compositional complexity of mixed stands, rather than by limitations specific to either input representation.

Despite the overall strong performance, classification accuracy remains lower for several species groups, including other broadleaves, maple, and larch in the pure forest category, and other broadleaves, birch, larch, and maple in the mixed category. 
These groups correspond to the least represented classes in the training data (Table~\ref{tab:distribution_labels}), suggesting that the scarcity of labeled samples may have contributed significantly to their reduced performance. 
This imbalance reflects the actual distribution of forest tree species across Denmark that is dominated by Norway spruce, European beech, and pedunculate oak \citep{NordLarsenEtAl2025}.
Future work could explore targeted data collection efforts for underrepresented species or investigate the use of synthetic oversampling techniques and cost-sensitive learning approaches to improve model performance in minority classes \citep{chawla2002smote, mouret2025tree}.

\subsubsection{Trade-Off Between Accuracy and Operational Demands}

Despite the superior performance of STF, the performance gap relative to TESSERA is comparatively modest. 
This finding is notable given that TESSERA is not specifically designed for tree species mapping. 
Its competitive performance suggests that the embeddings retain transferable information relevant to species-level forest classification.

The two approaches, however, differ substantially in their operational demands. 
Constructing STF requires acquiring and processing multi-temporal Sentinel-1 and Sentinel-2 imagery through a computationally demanding workflow involving radiometric harmonization, cloud masking, co-registration, temporal interpolation, and spectral index calculation, a pipeline that is both resource-intensive and requires domain expertise. 
In contrast, FM embeddings are available as analysis-ready representations that can be used directly in downstream classification pipelines.
This significantly reduces preprocessing effort and lowers barriers to rapid implementation at large spatial scales.
Consequently, the relatively modest reduction in classification performance may represent an acceptable trade-off in operational settings where scalability, reproducibility, and deployment efficiency are prioritized alongside accuracy.
However, reliance on pre-computed embeddings also reduces user control over the input processing and may create dependencies on maintenance and the continued availability of externally produced embedding products.

An additional consideration is model interpretability.
The ablation analysis performed on STF provided insights into the relative contribution of different inputs and enabled a more transparent understanding of model behaviour. 
A comparable analysis is considerably more challenging for FM embeddings, where individual embedding dimensions encode a complex, entangled combination of information from multiple modalities without explicit physical correspondence.
Although significant progress has been made toward understanding FM representations, the internal mechanisms of these models remain largely opaque \citep{lu2025representation}. 
This lack of interpretability may limit their suitability in operational contexts where transparent and accountable decision-making is required, such as in regulatory or conservation planning frameworks.

As shown in Figure~\ref{fig:mlp_tessera_report}, the relative performance of the STF and TESSERA representations changes as the amount of training data decreases. Using training data fractions below approximately 25\% of the full training set, the TESSERA MLP consistently achieves higher macro F1 scores than the STF MLP, and the difference increases as fewer training samples are used.
This result suggests that FM embeddings encode transferable information that reduce the amount of labeled data required to learn effective class boundaries \cite{xiao2025foundation}.

The advantage of FM embeddings in low-data settings is particularly relevant for forest monitoring applications. 
Reference datasets are often constrained by fieldwork costs, limited accessibility, safety considerations, and uneven spatial coverage \citep{de2020evaluating, queinnec2022developing}.
This can result in training datasets that are limited in size, geographically biased or temporally outdated for developing highly specialized models \citep{he2026terrain}.
In such scenarios, the improved label efficiency of FM embeddings offers a meaningful operational advantage over STF approaches.

\subsubsection{Sensitivity to temporal input configuration}

Both STF and TESSERA benefit from the inclusion of multiple years, with the three-year configuration achieving the highest F1 score on both pure and mixed plots. This finding is consistent with those of Pflugmacher et al. \citep{Pflugmacher2019} who showed that temporally dynamic classes such as cropland and grasslands benefit the most from more frequent observations.
This can be attributed to several factors, such as variability in cloud cover, the gradual development of canopy structure in younger stands, whose spectral and structural signals strengthen as trees mature, and variability in forest phenology driven by climatic conditions \citep{piao2019plant}.

The two representations, however, differ in their sensitivity to the selected input years.
For pure plots, the STF MLP trained on 2022 data performs 10.0 percentage points below the full three-year model, while the 2021--2022 configuration performs 5.6 percentage points below it.
The weakest STF performance therefore occurs in configurations that include 2022 but exclude 2020. 

In contrast, TESSERA exhibits less variation among the individual-year configurations and outperforms STF for both the 2022 and 2021--2022 configurations on pure plots. For 2022, TESSERA achieves an F1 score of 0.793 compared with 0.743 for STF, while for 2021--2022 the corresponding scores are 0.809 and 0.787.
This suggests that TESSERA is less sensitive than STF to the selection of input years. 
One possible explanation is that the annual embeddings integrate information learned during large-scale pre-training and are therefore less dependent on the availability of a dense local optical time series. 
However, the present experiment does not isolate the effects of cloud cover, observation density, phenological variability, or pre-training, and broader robustness to year-specific data quality cannot be established from three years alone.

These findings add an important nuance to the overall comparison: while FM embeddings do not surpass manually engineered features in the full multi-year setting, their greater robustness to data quality variability represents a practically relevant advantage in operational monitoring contexts where complete, cloud-free time series cannot always be guaranteed.

\subsubsection{Influence of Pre-training Design on Downstream Performance of FMs}
 
Although both TESSERA and AlphaEarth provide general-purpose EO embeddings, the former consistently outperformed the latter across our experiments. 
This difference indicates that not all FMs are equally suited to fine-grained RS applications. 
Instead, downstream performance appears to be strongly influenced by model design and pre-training strategy. 

One possible explanation lies in the differing objectives used during model development. 
TESSERA was explicitly designed for pixel-level representation learning and trained to capture local spectral and temporal dynamics, characteristics that align closely with the pixel-based classification framework employed in this study.
AlphaEarth, while also producing spatially resolved pixel-level embeddings, was designed as a broader multi-source geospatial representation model that integrates spatial, temporal, and cross-sensor information. Its more heterogeneous pretraining framework therefore provides a complementary representation to TESSERA's explicitly pixel-wise spectral-temporal formulation.
The weaker performance observed here may therefore reflect a mismatch between pre-training objectives and the spatial granularity required for detailed tree species discrimination. 

These findings suggest that the suitability of EO FMs cannot be assessed independently of the downstream application. 
Rather, the alignment between pre-training strategy and target task characteristics may be a critical, yet still underexplored, determinant of downstream performance. 
A more systematic evaluation of how FM design influences performance across applications of varying spatial and thematic granularity remains an important direction for future research.

\subsection{Implications for Large-Scale Forest Monitoring}

\subsubsection{Treatment of Mixed Forest}
Mixed forest plots are often excluded in tree species mapping studies, although they can represent a significant proportion of the forested area. Building on recent efforts to include mixed plots in national-scale mapping evaluation \citep{blickensdorfer2024national, schadauer2024evaluating}, we incorporate mixed forest plots in model evaluation and in the accuracy assessment of the final map.

The lower performance observed for mixed plots can be attributed to two key factors. 
First, a single dominant-species label cannot fully represent the spectral and structural information of a plot that contains several species. The four central pixels may include different species or mixtures of crowns, especially where the dominant species accounts for only slightly more than half of the plot area.
Second, all models are trained exclusively on pure forest plots, where the dominant species occupies most of the canopy, making generalization to mixed plots more challenging.
The resulting performance reduction should consequently be interpreted as a combined effect of greater reference-label ambiguity, within-plot heterogeneity, and train--test distribution shift.

To account for this ambiguity, we evaluate predictions at the plot level using the aggregation rule described in Section~\ref{sec:validation}, which does not require every pixel to be correctly classified and is therefore better suited to the spatial heterogeneity of mixed plots. Nevertheless, classification performance remains substantially lower for mixed than for pure forest plots, highlighting the challenge of mapping compositionally heterogeneous forests from medium-resolution satellite data.
This also reflects a limitation associated with the NFI sampling design, which relies on concentric circular plots with different measurement radii depending on tree size. Species composition is therefore estimated by scaling tree measurements obtained from the different radii to the full plot area, introducing additional uncertainty when plot-level composition is linked to individual pixels. Furthermore, because spatial tree positions are available only for the permanent sample plots, which constitute approximately one-third of the plots, we refrain from using this information when matching plots to individual pixels. Collectively, potential mismatches related to tree scaling and positioning introduce additional uncertainty when linking plot-level labels to pixel-based predictions. This is particularly relevant in mixed plots, where species composition and spatial arrangement may vary substantially within the plot. Future research could investigate whether explicitly aligning the available tree positions with the satellite pixels improves pixel-level label assignment and reduces reference uncertainty.

\subsubsection{National-scale mapping and area estimation}
This study contributes to a growing number of works demonstrating the capability of Sentinel-based approaches for national-scale forest tree species mapping \citep{francini2024forest, wegler2025tree, blickensdorfer2024national, grabska2024map, breidenbach2021national}.
The final map extends spatially explicit species information beyond the publicly managed forests covered by the available stand-level polygon dataset and provides continuous dominant-species predictions across Denmark.

In addition to multi-temporal Sentinel-1 and Sentinel-2 imagery, we complement our input data with spectral indices and canopy height metrics.
The resulting performance highlights the value of combining complementary spectral, structural, and temporal information for large-scale forest mapping. In this respect, the study aligns with broader efforts to operationalize satellite-based forest monitoring \citep{fao2020fra}.

The map-level assessment highlights an important distinction between spatial classification and statistical area estimation. 
The mapped areas calculated directly from the final map show varying levels of agreement with the national forest statistics, with notable discrepancies for beech, spruce, fir, and maple.
After adjustment using the map-accuracy sample, the estimated areas move substantially closer to the inventory-based statistics for most species.
For example, the estimated area for maple now differs by just 1,071 ha, compared with a difference of 11,741 ha based on the mapped pixel counts. 
OBL is the only class for which the adjusted estimate is farther from the reference value than the direct map-based estimate.
This indicates that raw mapped pixel counts should not be interpreted as unbiased species-area estimates without accounting for omission and commission errors.

The resulting national forest species map demonstrates strong overall accuracy and is made publicly available, offering a valuable resource for operational forest monitoring and management by interested authorities. 
More broadly, the framework is not limited to a single static map product and could be adapted for repeated annual mapping, thereby supporting the monitoring of forest development and compositional change over time, with potential relevance for ecological assessment, carbon accounting, and management planning \citep{wulder2020satellite, fassnacht2024remote}.

\subsection{Limitations and Uncertainties}

Several limitations and sources of uncertainty should be acknowledged when interpreting the results of this study. 
A fundamental constraint arises from the nature of optical satellite observations, which capture spectral signals primarily from the uppermost canopy layer and are therefore unable to fully characterize subcanopy tree structure or composition. 
This is particularly relevant for tree species discrimination, where diagnostic differences between certain species may occur in lower canopy layers, in basal area distribution, or in structural attributes that are not directly observable from space. 
Furthermore, the spatial resolution of the Sentinel missions, while suitable for large-scale mapping, may be insufficient to resolve the fine-scale canopy heterogeneity that characterizes mixed forest stands and fragmented woodland patches. 
These constraints apply equally to FM embeddings, which are derived from the same satellite observations and are therefore prone to the same inherent sensing limitations.
A further limitation concerns the training setup itself. 
This study focuses exclusively on the dominant tree species within each inventory plot, without accounting for the full compositional complexity of co-occurring species. 
A multi-label classification formulation could, in principle, better reflect this compositional complexity. However, in the absence of intra-plot spatial information, many pixels would inevitably be assigned incorrect multi-label combinations, potentially introducing more noise than the current single-label approach.

Label noise also remains an inherent challenge in the current setup. 
This noise arises primarily from the spatial mismatch between pixel-level inputs and plot-level reference labels, and is particularly pronounced for mixed forest plots where multiple species coexist within a single reference unit. 
Although restricting classifier training to pure forest plots and aggregating pixel-level predictions at the plot level reduce the influence of this uncertainty, residual label noise inevitably remains. The application of dedicated label-noise learning approaches \citep{kondylatos2025probabilistic} represents a promising direction for future work.

Although the four pixels associated with each NFI plot are retained within the same data partition, the random plot-level split does not explicitly enforce geographic separation between the training and test sets. The study also does not assess temporal transfer to mapping years beyond 2022 without model retraining. Future work should therefore assess spatially blocked and cross-region validation, temporal transfer to subsequent years, and methods designed to account explicitly for uncertain or compositional labels.

\section{Conclusion}
This study developed and evaluated a national-scale tree species mapping framework for Denmark using NFI observations, Sentinel-1 and Sentinel-2 time series, and canopy height data. 
We compared conventional STF with embeddings derived from the EO FMs TESSERA and AlphaEarth across multiple ML classifiers.

The STF MLP achieved the highest overall performance. 
Nevertheless, TESSERA embeddings provided competitive results, particularly for pure forest stands, despite requiring substantially less preprocessing and no task-specific feature engineering. 
TESSERA also demonstrated greater robustness under limited training data conditions, outperforming the STF when fewer than 25\% of the available reference data were used for training. 
In contrast, AlphaEarth embeddings were consistently less effective for this application.

The results further demonstrated the importance of combining complementary sources of information for tree species discrimination.
Multi-year observations consistently improved classification performance relative to single-year inputs, while ablation analyses highlighted the contribution of the Sentinel-1 SAR data, canopy height metrics, and spectral indices.
Evaluations conducted separately for pure and mixed forest stands also revealed the increased difficulty of classifying mixed forests, underlining the importance of considering forest compositional complexity when evaluating operational tree species mapping approaches.

Using the best-performing STF MLP, we produced the first open-access 10 m national tree species map of Denmark. 
Area-adjusted validation yields an overall accuracy of 79.9\%, indicating that the map provides a reliable representation of dominant tree species patterns at national scale. 
The map is available here: \url{https://zenodo.org/records/22108850}.

Overall, our findings show that dense spectral-temporal Sentinel representations remain an effective approach for maximizing classification accuracy in national-scale tree species mapping. 
At the same time, the strong performance of TESSERA, particularly under limited training data availability, highlights the potential of FM embeddings as an alternative to traditional feature-engineering workflows. 
As EO FMs continue to mature, they have the potential to reduce reliance on task-specific feature engineering while supporting scalable and transferable approaches to large-scale forest monitoring.

\section*{CRediT authorship contribution statement}
\textbf{Alkiviadis Koukos:} Writing – original draft, Conceptualization,  Methodology, Software, Validation, Formal analysis, Investigation, Data curation, Visualization.
\textbf{Spyros Kondylatos:} Writing – review \& editing, Conceptualization, Methodology, Software, Validation, Formal analysis, Investigation, Visualization.
\textbf{Thomas Nord-Larsen:} Writing – review \& editing, Resources, Validation, Data Curation.
\textbf{Lotte Nyborg:} Validation, Resources, Funding acquisition, Project administration.
\textbf{Christian Tøttrup:} Funding acquisition, Project administration.
\textbf{Kenneth Grogan:} Writing – review \& editing, Conceptualization, Methodology, Validation, Formal analysis, Investigation, Visualization, Supervision.

\section*{Data availability.}
The map of the dominant tree species in Denmark is available \href{https://zenodo.org/records/22108850}{here}.

\section*{Declaration of competing interest}
The authors declare that they have no known competing financial interests or personal relationships that could have appeared to influence the work reported in this paper.

\section*{Declaration of generative AI and AI-assisted technologies in the manuscript preparation process.}
During the preparation of this work, the authors used Claude and Copilot to improve readability. After using these tools, the authors reviewed and edited the content as needed and take full responsibility for the content of the published article.

\section*{Acknowledgements}
This work was funded by DHI's research contract with the Danish Ministry of Higher Education and Science, Denmark (contract 5255-00005B) and by Innovation Fund Denmark through the INNO-CCUS partnership, as part of the INNO4EST project (Project No. 2-P3).

\bibliographystyle{elsarticle-num}

\end{document}